\documentclass{article}
\usepackage{arxiv}

\usepackage[utf8]{inputenc} 
\usepackage[T1]{fontenc}    
\usepackage{url}            
\usepackage{booktabs}       
\usepackage{amsfonts}       
\usepackage{nicefrac}       
\usepackage{microtype}      
\usepackage{lipsum}
\usepackage{graphicx}
\usepackage{amsmath}
\usepackage{bbm}
\usepackage[ruled,vlined]{algorithm2e}
\usepackage{float}
\usepackage{epstopdf}
\usepackage{parskip}
\usepackage{latexsym,amssymb}
\usepackage{caption}
\usepackage{subcaption}
\usepackage{amsthm}
\usepackage{natbib}

\usepackage[english]{babel}
\usepackage[colorlinks=true,linkcolor=red,filecolor=green,citecolor=blue]{hyperref}
\usepackage[dvipsnames]{xcolor} 
\usepackage{cleveref}

\author{%
  Stefanos Bennett\thanks{corresponding author}
  \\
  Department of Statistics\\
  University of Oxford \& \\
  The Alan Turing Institute, London, UK \\
  \texttt{stefanos.bennett@stats.ox.ac.uk}  \\ 
    \and
  \textbf{Mihai Cucuringu}\\
  Department of Statistics \& Mathematical Institute\\
  University of Oxford \&  \\
  The Alan Turing Institute,  London, UK \\ 
  \texttt{mihai.cucuringu@stats.ox.ac.uk} \\
    \and
 \textbf{Gesine Reinert} \\ 
  Department of Statistics\\
  University of Oxford  \&  \\ 
  The Alan Turing Institute,  London, UK \\
  \texttt{reinert@stats.ox.ac.uk} \\
}

\title{Lead-lag detection and network clustering for multivariate time series with an application to the \\ US equity market}

\begin{document}

\maketitle

\begin{abstract}
In multivariate time series systems, it has been observed that certain groups of variables partially lead the evolution of the system, while other variables follow this evolution with a time delay; the result is a lead-lag structure amongst the time series variables. In this paper, we propose a method for the detection of lead-lag clusters of time series in multivariate systems. We demonstrate that the web of pairwise lead-lag relationships between time series can be helpfully construed as a directed network, for which there exist suitable algorithms for the detection of pairs of lead-lag clusters with high pairwise imbalance. Within our framework, we consider a number of choices for the pairwise lead-lag metric and directed network clustering components. Our framework is validated on both a synthetic generative model for multivariate lead-lag time series systems and daily real-world US equity prices data. We showcase that our method is able to detect statistically significant lead-lag clusters in the US equity market. We study the nature of these clusters in the context of the empirical finance literature on lead-lag relations and demonstrate how these can be used for the construction of predictive financial signals.
\end{abstract}


{
}

\section{Introduction}

Multivariate time series are ubiquitous in a wide range of domains, such as the physical sciences, medicine, and economics. Often, multivariate systems describing multiple processes or quantities are thought to exhibit lead-lag relationships \citep{Podobnik2010a}. In this work, time series A is said to lead time series B if A's past values are more strongly associated with B's future values than A's future values are with B's past values. The study of lead-lag relationships in multivariate time series systems is of interest in fields such as earth science \citep{Harzallah1997}, biology \citep{Runge2017} and economics \citep{Wang2017, Sornette2005}. For example, \cite{Harzallah1997} study the lead-lag relationship between the Indian summer monsoon and a number of climate variables such as snow cover, sea surface temperature and geopotential height across a grid of locations on the Earth's surface. \cite{Wang2017} examine the lead-lag dependence between the spot and futures markets for a Chinese stock market index.

In this paper, we examine systems of lead-lag relationships in time series data through the lens of directed network analysis. By constructing a network based on pairwise lead-lag metrics between variables, we are able to study overall properties of the web of lead-lag relationships via the tools of network analysis. Our specific interest lies in discovering clusters of different variables that exhibit strong lead-lag behaviour. To this end, we employ unsupervised directed network clustering and leverage recently developed algorithms  \citep{Cucuringu2019} that identify clusters with high imbalance in the flow of weighted edges between pairs of clusters.

While we expect our unsupervised learning method to be applicable to a number of multivariate time series domains, the particular application domain of interest in this study is the analysis of lead-lag clusters in financial time series data. Large financial markets, such as the US equity market, exhibit complex non-linear behaviour, often with a low signal-to-noise ratio \citep{Cont2001}.
By using pairwise lead-lag detection and network analysis tools, we aim to extract clusters that capture the latent lead-lag relationships which may be present in such complex systems. Furthermore, persistent historical clusterings can be utilised for the challenging task of returns forecasting. As a result, our unsupervised learning method may prove to be a valuable component in certain financial forecasting pipelines. Beyond financial markets, this approach may lead to insights into the nature of lead-lag relationships in climate \citep{Harzallah1997}, social \citep{Lin2013LeaderfollowerFV},  biological \citep{Runge2017} or economic systems \citep{iyetomi2020relationship,CAMILLERI2019170}.

\paragraph{Problem description} \label{sec:Problem setting}

In the context of multivariate time series systems, the problem of lead-lag detection consists in identifying random variables that lead or lag other random variables.  
Here, a variable $\alpha$ is said to {\it lead} another variable $\beta$ if $\alpha$'s past values are more strongly associated with $\beta$'s future values than $\alpha$'s future values are with $\beta$'s past values. In this situation, we also say that $\beta$ {\it lags} $\alpha$. 
There are a number of ways to mathematically define and extract the pairwise relationship between time series. Different lead-lag definitions are compared using a-priori considerations in \Cref{subsec:Pairwise metric of lead-lag relationship} and synthetic experiments in \Cref{sec:Synthetic data experiment}.

Once we have chosen a metric to capture lead-lag relations, we can represent the uncovered relations using a directed weighted network. The nodes of our network correspond to different time series variables. A directed edge $\alpha \rightarrow \beta$ exists between nodes $\alpha$ and $\beta$ if $\alpha$ leads $\beta$. The weight of this edge is given by the magnitude of the pairwise lead-lag metric, thus encoding the strength of the relation. We are thus able to study the properties of lead-lag relationships using the tools of network analysis. 

A key question in network analysis concerns community detection \citep{newman2018networks}. Does there exist a clustering of nodes such that node similarity is, on average, stronger within clusters than between clusters? In the context of a directed network encoding  lead-lag relations, the question of community detection can be  
framed in terms of identifying clusters that exhibit high pairwise cut imbalance, as follows. 
We regard the flow along a directed weighted edge $\alpha \rightarrow \beta$ as a measure of the extent to which  $\alpha$ leads $\beta$. In a directed graph with adjacency matrix $A$, the cut associated to two subsets of nodes $\mathcal{A}$ and $\mathcal{B}$, is given by $Cut(\mathcal{A},\mathcal{B}) = \sum_{i\in \mathcal{A}, j \in \mathcal{B}} A_{ij}$, and we refer to the difference $ Cut(\mathcal{A},\mathcal{B}) - Cut(\mathcal{B}, \mathcal{A})$ as the {\it cut imbalance}. A high cut imbalance between communities $\mathcal{A}$ and $\mathcal{B}$
indicates that variables in $\mathcal{A}$ are, on average, leaders of variables in $\mathcal{B}$. Therefore, by identifying pairs of clusters with high imbalance, we segment our multivariate system into communities that, taken in pairs, are mostly composed of either leaders or laggers. In  \Cref{sec:Method}, we describe a Hermitian-based directed network clustering algorithm that is suited for this task following \citep{Cucuringu2019}.

The application domain studied in this paper is that of financial time series. In this domain, each time series corresponds to the return time series for a particular financial instrument. We investigate the lead-lag cluster structure of the US equity market. In particular, we are interested in four questions. 
Does there exist a statistically significant cluster structure in the {US equity market}? What is the nature of the data-driven clustering? How does the data-driven cluster structure relate to previously discovered lead-lag mechanisms? Can we leverage our clustering for downstream  forecasting purposes?

\paragraph{Key contributions}
Our primary contribution is the introduction of a principled method, which, to the best of our knowledge, is the first to address the problem of unsupervised clustering of leading and lagging variables in multivariate time series systems. We validate different components of our method on synthetic and real data sets. Our secondary contribution consists of an evaluation of novel pairwise lead-lag metrics using a new benchmark data generating process for multivariate time series systems with clustered lead-lag structure. Thirdly, the application of our method to US equity data provides insights into the structure of the US equity market. To the best of our knowledge, our work presents the first data-driven clustering of lead-lag networks in a financial market context. Finally, we construct a novel statistically significant trading signal for the US equity market --  thus demonstrating how our method can be employed to extract valuable signals in the high-dimensional, low signal-to-noise data setting.

\paragraph{Paper outline}
We discuss existing literature related to our work in Section \ref{sec:Related work}. Section \ref{sec:Method} describes our approach to solving the lead-lag extraction and clustering problems. In Section \ref{sec:Synthetic data experiment}, we validate our method on synthetic data sets. We present the results of applying our algorithm to a universe of US equities in Section \ref{sec:US equity data experiment}. In \Cref{sec:Financial forecasting application}, we illustrate the use of our methodology in a financial forecasting application. Finally, we summarise our main findings in \Cref{sec:Conclusion}.

\section{Related work} \label{sec:Related work}

There exists substantial evidence of lead-lag relations at the scale of monthly, weekly and daily financial returns \citep{Lo1990, Badrinath1995, Brennan1993, Chordia2000, Menzly2010, Cohen2008}, as well as at higher frequencies \citep{Huth2012, Wang2017, Curme2015a, Curme2015}. In addition, a number of studies have considered lead-lag relations from the point of view of networks \citep{Curme2015, Fiedor2014, Vyrost2015, Liao2014, Sandoval2014a, Billio2012, Wang2017a, Wu2010}. Commonly studied questions in this financial lead-lag network literature concern the cluster structure of the lead-lag network \citep{Sandoval2014a, Billio2012, Liao2014, Wang2017a, Xia2018, Biely2008}. A number of papers consider the relative influence of different industry sectors within the lead-lag network \citep{Biely2008, Liao2014, Xia2018}. The influence of various sub-sectors within the lead-lag network of financial institutions is also a particular question of concern \citep{Billio2012, Wang2017a, Sandoval2014a}. For example, \cite{Billio2012} relate the lead-lag network to the systemic exposure of financial firms and sub-sectors,  in order to understand their respective financial drawdowns during crisis periods. In addition, the effect of geography-based clusters has also been investigated \citep{Sandoval2014a}.

A second commonly studied problem in the financial lead-lag literature is that of ranking. A number of lead-lag network papers focus on how network tools may be used to identify financial instruments that exhibit stronger tendencies to lead other instruments \citep{Liao2014, Billio2012, Wu2010, Basnarkov2019, Stavroglou2017}. For example, \cite{Wu2010} and \cite{Basnarkov2019}, apply the PageRank algorithm \citep{Kumar2012} to the lead-lag network in order to extract an ordering of equities in terms of their influence on the future values of other equities.

In addition to the literature on financial \textit{lead-lag} correlation networks, there is also substantial literature on \textit{synchronous} correlation networks \citep{Tumminello2010, Namaki2011a, Sandoval2012, Marti2017}. The reader is referred to \cite{Marti2017} for an extensive review of clustering on (mostly) synchronous financial correlation networks.

Our empirical analysis is novel within the financial lead-lag literature since it is the first work to extract a data-driven clustering of the lead-lag network. In contrast, previous studies \citep{Sandoval2014a, Billio2012, Liao2014, Sandoval2014a, Wang2017a, Xia2018, Biely2008} are only able to capture the influence of predefined groups, which are given, for instance, by industry sector \citep{Biely2008, Liao2014, Xia2018} or geography \citep{Sandoval2014a}, within the financial lead-lag network. We believe that the academic interest in our data-driven clustering approach is underscored by the plurality of papers \citep{Marti2017} that apply data-driven clustering to 
synchronous correlation networks, as well as the number of papers that apply data-driven ranking methods to 
lead-lag networks \citep{Liao2014, Billio2012, Wu2010, Basnarkov2019, Stavroglou2017}. Furthermore, our work is the first to show that 
clustered lead-lag network structure can be successfully used for downstream out-of-sample prediction tasks.

\section{Method} \label{sec:Method}
Our method is a pipeline consisting of three steps. First, we apply a pairwise lead-lag metric to capture the lead-lag relationship between each pair of time series; this results in a network of lead-lag relationships. Second, we apply a directed network clustering method to extract a partition of the multivariate system such that there is a large flow imbalance (net sum of weights of inter-cluster edges \citep{Cucuringu2019}) between cluster pairs. 
The third step quantifies the \textit{leadingness} of each cluster. 

There are a number of choices for each of these components in our pipeline. In this section, we describe metrics that can be used to quantify lead-lag relations between pairs of time series, and available directed network clustering methods.

To introduce notation, let $X^i_t$ denote the random value of the time series variable $i \in \{1, ..., p\}$ at time $t = 0, \ldots, T$. Further, define the first differences $Y^i_t = X^i_t - X^i_{t - 1}$ for $i \in \{1, ..., p\}$, $t = 0, \ldots, T$. 

In our application domain of US equities, $X^i_t$ denotes the logarithm of the closing price for stock $i \in \{0, 1, ..., p\}$ on day $t = 0, \ldots, T$. Hence $Y^i_t$ provides the corresponding log-return for equity $i$ from day $t - 1$ to $t$.
It is suitable to use log-returns for analysis as they exhibit closer to stationary properties, and log-returns are more mathematically tractable than linear or percentage returns in the computation of multi-horizon returns \citep{Campbell1997}.

\subsection{Pairwise metrics of lead-lag relationship} \label{subsec:Pairwise metric of lead-lag relationship}

In a complex, non-linear system such as the US stock market, determining a suitable way to define a metric to capture lead-lag relationships is challenging. Here we present some options.

\subsubsection{Lead-lag metrics based on a functional of the cross-correlation} \label{subsubsec:Lead-lag metrics based on a functional of the cross-correlation}

A commonly used approach to defining a lead-lag metric is to use a functional of a sample cross correlation function (ccf) between two time series. 
A {\it sample cross-correlation function} between time series
$i$ and $j$ evaluated at lag $l \in \mathbb{Z}$ is given by  
\begin{equation}  \label{eq:cross-correlation-function}
\mathrm{ccf}^{ij}(l) = \mathrm{corr}\left(\{Y^i_{t - l}\}, \{Y^j_t \}\right), 
\end{equation} 
where $\mathrm{corr}$ denotes a choice of sample correlation function. The corresponding {\it lead-lag metric}, a measure of the extent to which $i$ leads $j$, is then obtained by
\begin{equation} \label{eq:lead-lag-metric}
    S_{ij} = F(\mathrm{ccf}^{ij}), 
\end{equation}
where $F$ is a suitable functional. 

In this paper, we consider four choices for the sample correlation function $\mathrm{corr}$, namely Pearson linear correlation,  Kendall rank correlation \citep{Kendall1938}, distance correlation \citep{Szekely2007}, and mutual Information based on discretised time series values \citep{Fiedor2014}. 
The four different sample correlation functions are able to detect different dependencies. Pearson correlation is able to detect linear dependencies, Kendall rank correlation is able to detect monotonic non-linear dependencies, while distance correlation and mutual Information are able to detect general non-linear dependencies. The drawback of non-linear sample correlation functions is that they have lower power in the case of a true linear relationship. 

Further, we consider two choices for the functional $F$,  as follows
\begin{enumerate}
    \item \textbf{ccf-lag1}: computes the difference of the cross-correlation function at $lag \in \{-1, 1\}$ 
    $$S_{ij} = \mathrm{ccf}^{ij}(1) - \mathrm{ccf}^{ij}(-1),$$
    \item \textbf{ccf-auc}: computes the signed normalised area under the curve (auc) of the cross-correlation function     $$S_{ij} = \frac{\mathrm{sign}(I (i,j) -I(j,i))   \cdot \mathrm{max}(I (i,j), I(j,i))}{I (i,j)+ I(j,i)},$$
    where $I (i,j)  = \sum_{l = 1}^L \left\vert \mathrm{corr}\left(\{Y^i_{t - l}\}, \{Y^j_t \}\right) \right\vert$ for a user-specified maximum lag $L$. 
\end{enumerate}

The $\textbf{ccf-lag1}$ method used with Pearson correlation is a crude lead-lag indicator \citep{Campbell1997}. This lead-lag indicator is only designed to take into account positive cross-correlation. Indeed, like the signatures-based method described further  below in Subsection \ref{sub:sig}, it is only able to correctly determine the direction of the lead-lag relationship under a positive cross-correlation association between time series. Thus, this lead-lag indicator should be restricted to domains such as US equity returns, where cross-correlations between time series variables are predominantly positive \citep{Campbell1997}.

The $\textbf{ccf-auc}$ method accounts   for  both positive and negative associations across multiple lags $l \in \{-L, \ldots, L\}$. The maximum lag $L$ can be chosen a-priori as the maximum time lag expected in the multivariate system, or by using cross-validation on some downstream validation criterion. The averaging approach $\textbf{ccf-auc}$ presented here is similar to the lag aggregation methodology of \cite{Wu2010}\footnote{We have also considered using a maximum aggregation approach, and have found similar qualitative results to the averaging-based approach presented in this paper; however, the maximum aggregation approach tends to perform slightly worse than the averaging-based approach.}.

Overall, we consider eight possible choices for lead-lag metrics based on functionals of the cross-correlation. This stems from four possible choices for correlation (Pearson, Kendall, distance correlation and mutual information) and two possible choices for the functional form ($\textbf{ccf-lag1}$ and  $\textbf{ccf-auc}$).   

The functional cross-correlation approach is flexible and computationally simple. The flexibility of the framework permits the use of robust and non-linear correlation metrics. The use of such non-linear correlation metrics is particularly useful for the extraction of lead-lag relationships in the financial time series domain, where linear cross-correlations between returns are expected to be low. High information efficiency in US equity markets \citep{Malkiel1970} implies that linear return cross-correlations are too low to be used to construct trading systems that have expected returns in excess of market equilibrium expected returns. 
On the other hand, a stylised feature of financial returns is volatility clustering \citep{Cont2001}; the size of the cross-correlation between the volatility of returns is expected to be larger than the cross-correlation between the raw returns themselves. A linear cross-correlation approach is unable to capture the relationship between the volatility of two instruments across time. Empirical studies have also found that stronger lead-lag relationships can be detected when taking into account volatility \citep{Billio2012}. Thus, when comparing the time-dependence in returns between two assets, we should allow for non-linear effects \citep{Fiedor2014}. In addition, the functional cross-correlation approach easily permits the use of correlation metrics that are robust to outliers. Since financial times series exhibit heavy tails \citep{Cont2001}, robustness constitutes an important feature for a lead-lag extraction method. In general, the functional cross-correlation component and, consequently, the entire pipeline will be robust to outliers if the choice of correlation metric is robust to outliers. For example, ordinal association correlation metrics such as Kendall correlation guarantee robustness to outliers.

The linear Granger causality approach that is often considered in financial lead-lag studies \citep{Shojaie2021} can be viewed as an extension of a functional linear cross-correlation-based approach that takes into account auto-correlation and also filters for statistical significance. General Granger causality methods may also use non-linear functional forms to capture the association between time series. These more general methods can be used as the lead-lag extraction component of our method. For the purposes of demonstrating our method and using robust and non-linear lead-lag extraction, simpler functional cross-correlation approaches will suffice. Since the combination of data auto-correlation and co-movement can produce lead-lag associations between time series variables using our method, one must be careful not to interpret resulting lead-lag associations as apparent causal influence estimates.

We contrast our approach, which is based on correlation networks, with causality-based approaches that attempt the more difficult problems of recovering the casual network underlying a multivariate time series system \citep{Runge2017} and quantifying its causal influences \citep{Janzing2013}. For example, whereas \cite{Runge2017} attempt to estimate the causal network underlying the time-lagged dependency structure in a given multivariate time series system, our aim is estimating and clustering the association-network for the multivariate time series system. Association-based approaches are more common in the financial network lead-lag literature \citep{Marti2017}, since financial time series have very noisy returns and exhibit weak lead-lag effect sizes due to the informational efficiency of the market \citep{Malkiel1970}. These characteristics of financial returns make the problem of accurately estimating a lead-lag correlation network (let alone the causality network) challenging in itself.

\subsubsection{Lead-lag metric based on signatures}\label{sub:sig} 

The approach of using a functional of the cross-correlation function relies on the user to specify the choice of functional; this choice is not obvious in many cases. In particular, it is difficult to gauge the number of lags to incorporate into our lead-lag metric a-priori. An alternative approach draws on the idea of signatures from rough path theory \citep{Levin2013a}, in order to construct a pairwise lead-lag metric. The signature of a 
continuous path with finite 1-variation \citep{Levin2013a}
$X : [a, b] \rightarrow \mathbbm{R}^d$, denoted by $S(X)_{a,b}$, is the collection of all the iterated integrals of $X$, namely $S(X)_{a,b} = (1, S(X)^1_{a,b}, \ldots, S(X)^d_{a,b}, S(X)^{1,1}_{a,b}, S(X)^{1,2}_{a,b}, \ldots)$, where the iterated integrals are given by
$$
    S(X)_{a, t}^{i_{1}, \ldots, i_{k}}=\int_{a<t_{k}<t} \cdots \int_{a<t_{1}<t_{2}} d X_{t_{1}}^{i_{1}} \ldots d X_{t_{k}}^{i_{k}} .
$$
Based on the proposal in \cite{Levin2013a}, the signatures-based pairwise measure of the lead-lag relation between two stocks $i$ and $j$ over the time period $[t - m , t]$ is given 
by
\begin{equation}\label{eqn:sig_lead_lag_measure}
S_{ij}(t - m, t) = \iint \limits_{t-m < u < v < t} ( dX^i(u)dX^j(v) - dX^j(u)dX^i(v)  ) .
\end{equation} 
This is the difference in the cross-terms of the second level of the time series signature of the log-prices. Theoretical results in rough path theory \citep{Levin2013a} have established that a signature is essentially  unique to the path it describes, and that the truncated signature (i.e. the lower order terms) can efficiently describe the path. \cite{Chevyreva} provide an interpretation of the signature lead-lag metric \eqref{eqn:sig_lead_lag_measure}. The signature lead-lag metric is positive and grows larger whenever increases (resp. decreases) in $X^i$ are followed by increases (resp. decreases) in $X^j$. If the relative moves of $X^i$ and $X^j$ are in the opposite directions, then the signature lead-lag measure is negative. Note that a downside of this method is that it is not able to tell the difference between
\begin{enumerate}
    \item $i \rightarrow j$ with negative association, 
    \item $i \leftarrow j$ with positive association.
\end{enumerate}
As a result, we do not expect the method to perform well when there is significant negative association in the lead-lag data generating process.

When analysing price data observed at discrete time points, we transform the data stream into a piecewise linear continuous path and calculate the second order signatures \citep{Reizenstein2018}. From this, we may calculate the lead-lag relation using the difference in second order signature cross-terms \eqref{eqn:sig_lead_lag_measure}. 
We refer the reader to \cite{Gyurko2014} for additional details on signatures and their application in a financial context, along with an interpretation in terms of second order areas and interplay with lead-lag relationships.
In practice, when comparing the signature lead-lag metrics across different pairs of time series, we recommend the normalisation of the price data prior to computation of the lead-lag metric, since the absolute value of the metric is increasing in the volatility of the underlying price series.

\subsubsection{Alternative lead-lag metrics}

The lead-lag extraction approaches mentioned in this section are by no means  exhaustive. Indeed, alternative methods can be found within the financial time series lead-lag literature \citep{Wang2017}. Furthermore, the functional cross-correlation framework presented in this paper is agnostic to the choice of the  correlation metric used within it. As such, it is able to draw on a wide array of non-linear correlation metrics such as target/forget dependence coefficient \citep{Marti2016}, maximal information coefficient \citep{Reshef2011} or maximum mean discrepancy \citep{Gretton2012}.

\subsection{Algorithms for clustering directed networks} \label{subsec:Network clustering}
Let $S_{ij}$ denote the user-defined lead-lag metric that quantifies how much time series variable $i$ leads $j$. The value $S_{ij}$ can be positive or negative, and satisfies $S_{ij} = -S_{ji}$. 
The lead relationships between all pairs of time series is encoded by the asymmetric matrix $A_{ij} = \max(S_{ij}, 0)$.  We apply directed network clustering algorithms to the weighted and directed network $G$, where each node corresponds to a time series variable and the adjacency matrix is $A$. In this section, we present different relevant clustering methods for such directed networks.

Note that as a pre-processing step for any of the clustering methods described below, it is  possible to filter the pairwise measurements $S_{ij}$ when constructing the network $A$. For example, \cite{Curme2015} apply significance thresholding, whereby an edge exists between two nodes only if the corresponding lead-lag metric is sufficiently large in magnitude.

\subsubsection{Naive symmetrisation clustering}

Popular undirected network clustering methods, such as spectral clustering \citep{Shi2000}, cannot be immediately applied to directed networks, since directed networks with asymmetric adjacency matrices have complex spectra. Traditional approaches for directed network clustering have applied spectral analysis to a symmetrised version of the directed network adjacency matrix \citep{Sussman2012, Pentney2005}. We consider a commonly used naive symmetrisation-based directed clustering method as a baseline \citep{Satuluri2011}. This naive method applies a standard spectral clustering  algorithm \citep{Shi2000}  to the undirected network with adjacency matrix $\Tilde{A} = A + A^T$. In this paper, the spectral clustering algorithm applied to the derived undirected networks uses $k$-means clustering on a projection onto the first $k$ non-trivial eigenvectors of the random-walk normalised graph Laplacian (we drop the first eigenvector since  for connected networks it is always the unit vector). The value of $k$, corresponding to the desired number of clusters, is a hyperparameter of the algorithm.

\subsubsection{Bibliometric symmetrisation clustering}

Naive symmetrisation methods  produce a clustering that only takes into account edge density and not edge direction. As a result, they are unable to target clusterings with high pairwise flow imbalance between clusters. \cite{Satuluri2011} propose the degree-discounted bibliometric symmetrisation that is able to take into account edge direction information. In the degree-discounted bibliometric symmetrisation, spectral clustering is applied to the adjacency matrix
$$
    \Tilde{A}=D_{o}^{-1 / 2} A D_{i}^{-1 / 2} A^{T} D_{o}^{-1 / 2}+D_{i}^{-1 / 2} A^{T} D_{o}^{-1 / 2} A D_{i}^{-1 / 2}, 
$$
where $D_i$ is the diagonal matrix of weighted in-degrees and $D_o$ is the diagonal matrix of weighted out-degrees. The degree-discounted bibliometric symmetrisation applies degree-discounting to a symmetrisation that sums the number of common in- and out- links between two pairs of nodes. Therefore,  clusters produced by this method are expected to group together nodes that have a relatively large number of parent (sender) and children (receiver) nodes in common \citep{Satuluri2011}. Degree-discounting is a technique that has been found to work well in tasks on graphs with highly skewed degree distributions.

\subsubsection{DI-SIM co-clustering}

\cite{Rohe2016} propose a co-clustering algorithm for directed networks. The co-clustering algorithm first computes a regularised graph Laplacian using $A$; this initial step is performed so that the  algorithm may deal with heterogeneous and sparse data. Then, co-clustering is performed by applying $k$-means on the $k$-largest of each of the left and right normalised singular vectors of the Laplacian. This co-clustering identifies two  partitions of nodes: one partition groups together vertices with similar sending behaviour, while the other partition groups together vertices with similar receiving behaviour. In this paper, we denote the clustering obtained using the left singular vectors as \textbf{DI-SIM-L} and the clustering obtained using the right singular vectors as \textbf{DI-SIM-R}. We consider both choices of clustering in our experiments.

\subsubsection{Hermitian clustering}

The Hermitian clustering procedure \citep{Cucuringu2019} for clustering directed networks considers the spectrum of the complex matrix $\Tilde{A} \in \mathbb{C}^{p \times p}$,  which is derived from the directed network adjacency matrix as $\Tilde{A} = i(A - A^T)$. Since $\Tilde{A}$ is Hermitian, it has $p$ real-valued eigenvalues which we order by magnitude $\vert \lambda_1 \vert  \geq \ldots \geq \vert \lambda_p \vert$. The eigenvector associated with $\lambda_j$ is denoted by $g_j \in \mathbb{C} ^ p$ where, in Euclidean norm, $\parallel g_j \parallel = 1$ for $1 \leq j \leq p$.

Algorithm \ref{algo:herm_clustering} describes the procedure for clustering the directed network $G$. In our implementation, we set the number of top eigenvectors used to $l = k$. 

\begin{algorithm}[H]
\textbf{Input}: A directed graph $G = (V, E)$ with Hermitian adjacency matrix $\Tilde{A}$; number of clusters $k \geq 2$; $\epsilon > 0$
\begin{enumerate}
\item Compute all the eigenvalue/eigenvector pairs of $\Tilde{A}$  \\ $\{(\lambda_1, g_1), (\lambda_2, g_2), \ldots, (\lambda_l, g_l) \}$  satisfying $\parallel g_j \parallel = 1$ and $\vert \lambda_j \vert > \epsilon$ $, \, \forall j \in \{1, \ldots, l \}$
\item $P \leftarrow \sum_{j=1}^l g_j g_j^T$
\item Apply a $k$-means algorithm with input rows of $P$
\item Return a partition of $V$ corresponding to the output of $k$-means
\end{enumerate}
\caption{Hermitian clustering algorithm.}
\label{algo:herm_clustering} 
\end{algorithm}

Note that in practice, for scalability purposes, one can bypass the computation of the entire $n \times n$ matrix $P$ in order to directly cluster using the embedding given by the top $l$ eigenvectors.

\cite{Cucuringu2019} study the performance of the algorithm theoretically and experimentally under data generated from a directed version of a stochastic block model that embeds latent structure in terms of flow imbalance between clusters. They show that the algorithm is able to discover cluster structures based on directed edge imbalance. This contrasts with previous spectral clustering methods that detect clusters based purely on the  edge-density of symmetrised networks. The Hermitian clustering algorithm is particularly suited to our setting of clustering lead-lag networks, since we aim to extract pairs of clusters with high flow imbalance. In addition, as a pre-processing step for this algorithm, we apply random-walk normalisation to the adjacency matrix $\Tilde{A}$, so that the method is robust to heterogeneous degree distributions \citep{Cucuringu2019}; we refer to the resulting algorithm as the \textbf{Hermitian RW} algorithm.

\subsection{Alternative clustering algorithms}

State-of-the-art modularity clustering algorithms such as the Leiden algorithm \citep{Traag2019} may be used on directed graphs using a directed modularity metric \citep{Dugue2015}. \cite{Dugue2015} optimise a modularity metric that compares the number of edges within clusters to the expected number of edges under a null model. However, such modularity-based algorithms, which return clusters based on edge density, are not suited to our goal of uncovering clusters of leading and lagging variables based on flow imbalance between clusters.

An adaptation of the Hermitian clustering method has been proposed in \cite{Laenen2020}. The method presented in \cite{Laenen2020} aims to discover a clustering that maximises a normalised flow metric between communities using the spectrum of a normalised Hermitian Laplacian matrix. As such, it could be used as an alternative to the Hermitian RW clustering algorithm considered in this paper. Also recently, \cite{MotifSpectralDirectedClustering_ANS_2020} proposed an algorithm for clustering weighted directed networks that employs motif-based spectral clustering to uncover flow imbalance relationships between pairs of clusters.  

Lastly, we draw attention to a recent approach introduced in \cite{he2021digrac} that extends the Hermitian-based clustering algorithm \citep{Cucuringu2019}. This recent method departs from standard approaches in the literature, and treats edge directionality not as a nuisance but rather as the main latent signal. It does so by introducing a graph neural network framework for obtaining node embeddings for directed networks in a self-supervised manner, while accounting for node-level covariates.

\subsection{The leadingness metric}
We introduce the concept of a \textit{meta-flow graph} in order to capture the aggregate weighted flow between pairs of clusters. The total \textit{flow} between any two clusters is given by the net of the normalised weights between all edges directed from one cluster to another. The skew-symmetric matrix that encodes this information is dubbed the \textit{meta-flow} matrix, which we denote by $F$. Mathematically
$$
F_{ij} = \frac{1}{\vert C_i \vert \, \vert C_j \vert}\sum_{l \in C_i, m \in C_j} \left[ A_{lm} - A_{ml} \right], 
$$
where $C_a$ denotes the set of all nodes in cluster $a \in \{1, \ldots, k\}$, and $i, j \in \{1, \ldots, k\}, \, i \neq j$.  The diagonal of $F$ consists of zeros: $F_{ii} = 0, \, \forall i \in {1, \ldots, k}$.
We also define a metric for the \textit{leadingness}  of each cluster $i \in \{1, \ldots, k\}$, 
\begin{align} \label{eqn:cluster_leadingness}
  L(i):=  &\frac{1}{\vert C_i \vert} \sum_{{l \in C_i,}{m \in \{1, \ldots, p\}}} \left[ A_{lm} - A_{ml} \right].
\end{align} 
Thus, $L(i)$ averages the row-sums of the  skew-symmetric matrix $A - A^T$ for nodes within the cluster $i$; the row-sums of the lead-lag matrix provide a measure of the total tendency of the equity corresponding to the row to be a leader \citep{Huber1962}. From this metric, we obtain a ranking of the clusters from the most leading cluster (largest row-sum value), which we will label 0, to the most lagging cluster (smallest row-sum value), which has the largest numeric label $k-1$. In this paper, all data-driven clustering results will be presented using this labelling. 
The \textsc{RowSum Ranking} \citep{Huber1962, gleich2011rank} algorithm is an instance of a ranking method that recovers a latent ordering of variables given variable pairwise comparisons. There exists a rich literature on ranking from pairwise comparisons. The goal in this literature is to infer the strength $\ell_i, i=1,\ldots,p$ or ranking of $p$ items given a (potentially incomplete) set of pairwise comparisons which encode a noise proxy for $\ell_i - \ell_j$. Alternative ranking algorithms that could be employed for defining the leadingness of a cluster include \citep{fogel2016spectral,syncRank,CaterinaDeBacco_Ranking,SVDRankSync_JMLR,BradleyTerry1952,Pageetal98}, as well as \cite{chau2020spectral} for rankings that incorporate any available node level covariates.

\subsection{Algorithmic complexity of the method}

Let us denote by $ \psi $ the cost of the pairwise lead-lag metric of choice. The cost of the lead-lag network construction step amounts to $O(p ^ 2 \psi)$, where $p$ is the number of time series. For example, for the linear Pearson correlation $O(\psi) = O(TL)$, where $L$ is the number of lags, and $T$ is the sample size. The cost of a spectral clustering algorithm for $k$ clusters, such as Hermitian clustering \citep{Cucuringu2019}, is $O(k  p^2) < O(p^3)$. Therefore, the overall complexity amounts to $O(p^2  T  L + k  p^2)$. 

In the large $p$ setting, the above pipeline can become computationally prohibitive. 
One approach to alleviate this amounts to subsampling $m$ pairs of time series
out of the ${p \choose 2}$ choices. This will lead to a comparison lead-lag matrix with only $m$ nonzero entries; for example, the choice of sampling each edge with probability $ \frac{\log p}{p} $ renders $m = O(p \log p)$.  Since computing the leading eigenvectors of a sparse matrix via an iterative power method-based approach can be performed in a running time that is linear in the number of nonzero entries in the matrix, this step takes $O(p \log p)$ time. Thus, the approximate method is almost linear in the number of edges in the comparison graph. If the underlying pairwise comparison graph is weakly connected, which in practice will be the case because correlations will not be zero, then a choice of sampling probability of $O(\frac{\log p}{p})$ results in a pairwise comparison graph that is weakly connected with high probability. In such a situation we would expect the clustering in the sampled network to be a reasonable reflection of the clustering in the true network.  
We refer the reader to 
\cite{graph_sparsification} for spectral algorithms and theoretical considerations of the closely related graph sparsification literature, and to \cite{Graph_Sampling_Hu} for a survey and taxonomy of graph sampling techniques.

\section{Synthetic data experiments} \label{sec:Synthetic data experiment}
The purpose of this section is to validate our method on synthetic experiments in which the ground truth lead-lag relationships and clusters are known. This approach  will also give an indication of the relative performance of each of our lead-lag metrics and clustering components, under different data generating settings.

\subsection{Synthetic data generating process}

We introduce five different lagged latent variable synthetic generating processes to test our method. The general form of these synthetic generating processes is a latent variable model whereby the lagged dependence on the latent variable $z$ induces the clustering amongst the different times series $\{y_t^i\}$. Mathematically, the synthetic data generating processes take the form
\begin{equation}
    \begin{split} \label{dataform} 
        z_t &\overset{i.i.d.}{\sim} F_z \, \forall t \in \{1, \ldots, T\}, \,  \quad 
        z_t:=0 \, \forall t \leq 0,  \\
        y_t^i &= g_{l_i}(z_{t - l_i}) + \epsilon_t^i, \, \quad  \epsilon_t^i \overset{i.i.d.}{\sim} N(0, \sigma_{\epsilon}^2) \, \forall t \in \{1, \ldots, T\}, \, i \in \{1, \ldots, p\}, 
    \end{split}
\end{equation}
    where the lag corresponding to time series variable $i$ is $l_i \in L$ and $L$ is the set of lag values. The choice of the shared latent variable distribution $F_z$ and the functional dependencies $g_{l}, l \in L$ on the latent variable $z$ determines the data generating process. 
    The factor-based form of the synthetic data generation is motivated by our application to US equities \citep{Fama1993, Jegadeesh1995}. For instance, early work by \cite{Jegadeesh1995} studies a lagged factor model in the context of lead-lag effects. See also \Cref{sec:US equity data experiment} for a discussion of hypothesised clustered lead-lag return structures in the US equity market. The synthetic data generating process considered in this section is a toy model that is designed to test whether our method can correctly detect and cluster time series in a factor-driven scenario.

    The five particular forms of \eqref{dataform} that we will consider are as follows. 
\begin{enumerate}
    \item \textbf{Linear}
        \begin{align}
            F_z &= N(0, 1) \mbox{ and }
            y_t^i = z_{t - l_i} + \epsilon_t^i.    
            \label{eq:synLinear}
        \end{align}
    \item \textbf{Cosine}
        \begin{align}
            F_z &= U(-\pi, \pi) \mbox{ and }
            y_t^i = \frac{1}{\sqrt{\pi}} \cos(l_i \cdot z_{t - l_i}) + \epsilon_t^i.
            \label{eq:synCosine}
        \end{align}
    \item \textbf{Legendre}
        \begin{align}
            F_z &= U(-1, 1) \mbox{ and }
            y_t^i = P^L_{l_i + 1} (z_{t - l_i}) + \epsilon_t^i. 
            \label{eq:synLegendre}
        \end{align}
    \item \textbf{Hermite}
        \begin{align}
            F_z &= N(0, 1) \mbox{ and }
            y_t^i = \frac{1}{\sqrt{l_i!}}P^H_{l_i + 1} (z_{t - l_i}) + \epsilon_t^i.  
            \label{eq:synHermite}
        \end{align}
    \item \textbf{Heterogeneous}
        \begin{align}
            z \in \mathbb{R}^K, \, F_z &= N_{K \times K}(0, I_{K \times K}) \mbox{ and }
            y_t^i = z^{f_i}_{t - l_i} + \epsilon_t^i. 
            \label{eq:synHetero}
        \end{align}
\end{enumerate} 
Here, $P^L_l$ and $P^H_l$ are respectively the Legendre polynomial of degree $l$ and the Hermite polynomial of degree $l$. In the heterogeneous case, the superscript $f_i \in \{1, \ldots, K\}$ indicates on which component of the multivariate factor $z$ the time series $i$ depends.

In these five data generating process scenarios, by design, the cross-covariance at lag $k \in \mathbb{N}$ between any two time series $i, j \in \{1, \ldots, p\}$ is 
$$\mathbb{E}\left[ ( y^i_{t - k} - \mathbb{E}[y^i_{t - k}] ) ( y^j_t - \mathbb{E}[y^j_t]) \right] = 0$$ whenever $k \neq l_j - l_i$ 
due to the independence of $z_t$ across time. In the linear data generating case, setting (1), when $k = l_j - l_i$, then we have that $\mathbb{E}\left[ ( y^i_{t - k} - \mathbb{E}[y^i_{t - k}] ) ( y^j_t - \mathbb{E}[y^j_t]) \right] =  \mathbb{E}[(z_{t - l_j}) ^ 2] \geq 0$. This induces a linear dependence between time series $i$ and time series $j$ through the single non-zero value in the cross-covariance function between these two time series. Considering the whole network of lead-lag relations, we find that $i \rightarrow j$ ($i$ is a leader of $j$) if and only if $l_i < l_j$. Since multiple time series share the same lag, this network is clustered: time series $i$ and $j$ share the same cluster if and only if $l_i = l_j$. Our synthetic experiments test our method's ability to correctly detect lead-lag relationships and recover the underlying ground-truth clustering structure of the lead-lag network.

The non-linear data generating settings (2) - (4) engender additional challenges for our lead-lag extraction method. Due to the respective orthogonality of the cosine functions $\{\cos(mx)\}_{m \in \mathbb{N}}$, Legendre polynomials $\{P^L_m(x)\}_{m \in \mathbb{N}}$ and Hermite polynomials $\{P^L_m(x)\}_{m \in \mathbb{N}}$, the linear cross-covariance evaluated at lag $k$ between two time series $i$ and $j$ is zero even when $k = l_j - l_i$. Thus we expect metrics based on linear or cross-covariance methods to perform poorly in these settings. 
Non-linear lead-lag metrics are required in order to detect a non-linear dependence of time series $j$ on time series $i$ at lag $k = l_j - l_i$. 

The heterogeneous data generating process setting adds a further independence condition on the relationship between two time series. In this case, the cross-covariance at lag $k$ is nonzero if and only if both $k = l_j - l_i$ and $f_i = f_j$ are satisfied. The additional factor component equality condition implies that time series $i$ and time series $j$ share the same cluster if and only if $l_i = l_j$ and $f_i = f_j$.

In our simulation studies,  we consider the performance of different configurations of our method as the noise level $\sigma$ of our idiosyncratic error increases. The following experiment parameter choices are considered:

\begin{itemize}
    \item Number of data points per time series: $T = 250$
    \item Number of time series: $p=100$
    \item The standard deviation of the idiosyncratic noise: $\sigma \in \{ 0, 0.2, 0.4, 0.6, 0.8, 1, 2, 3, 4\}$
    \item Latent variable lag dependence for each time series by experiment setting:
        \begin{itemize}
            \item Linear: $l_i = \lfloor \frac{i - 1}{10} \rfloor$ for $i = 1, \ldots, 100$
            \item Cosine: $l_i = \lfloor \frac{i - 1}{10} \rfloor + 1$ for $i = 1, \ldots, 100$
            \item Legendre and Hermite: $l_i = \lfloor \frac{i - 1}{10} \rfloor + 2$ for $i = 1, \ldots, 100$
            \item Heterogeneous: $f_i = \lfloor \frac{i - 1}{50} \rfloor$ for $i = 1, \ldots, 100$ while $l_i = \lfloor \frac{i - 1}{5} \rfloor$ for $i = 1, \ldots, 50$ and $l_i = \lfloor \frac{i - 51}{5} \rfloor$ for $i = 51, \ldots, 100$.
        \end{itemize}
\end{itemize}

The lag and factor structure implies that there are 10 clusters in the Linear, Cosine, Legendre and Hermite settings,  while in the Heterogeneous setting there are 20 clusters. In each configuration of our method, we set the clustering algorithm hyperparameter corresponding to the number of clusters to be equal to the ground truth number of clusters. The remaining hyperparameter choices for the different method configuration components are:

\begin{itemize}
    \item \textbf{ccf-auc}: the maximum cross-covariance lag: $L = 5$
    \item \textbf{DI-SIM co-clustering}: the regularisation parameter is set equal to the average row sum of the adjacency matrix \citep{Rohe2016} and the number of singular vectors used in the co-clustering is set equal to the ground truth number of clusters in each synthetic data generating setting.
    \item \textbf{Naive, Bibliometric and Hermitian RW clustering}: the number of eigenvectors used in the respective spectral clustering projections is set equal to the ground truth number of clusters.
\end{itemize}

\subsection{Performance metrics}
We employ different performance criteria to evaluate both components -- the lead-lag detection component and the clustering component -- of our method. In order to evaluate the lead-lag detection component, we calculate the proportion of correctly classified edges in the true underlying lead-lag network (i.e. the accuracy of correctly classifying the direction of the lead-lag relationship between two time series). In order to evaluate the clustering component, we calculate the Adjusted Rand Index (ARI) between the ground-truth clustering and the clustering recovered by our method. 

\subsection{Results}

\subsubsection{Marginal results over lead-lag extraction and clustering}

We present the results for the lead-lag metric and clustering stages separately. In this section, we present results for the linear and cosine synthetic data generating settings; results for the other synthetic data generating settings can be found in Appendix sections \ref{sec:Appendix - Synthetic data experiment: lead-lag results} and \ref{sec:Appendix - Synthetic data experiment: clustering results}. For each experimental setting, we have generated 48 samples from the synthetic data generating process and applied our method to each one.

\begin{figure}[h!] 
	\centering
		\includegraphics[width=1\textwidth]{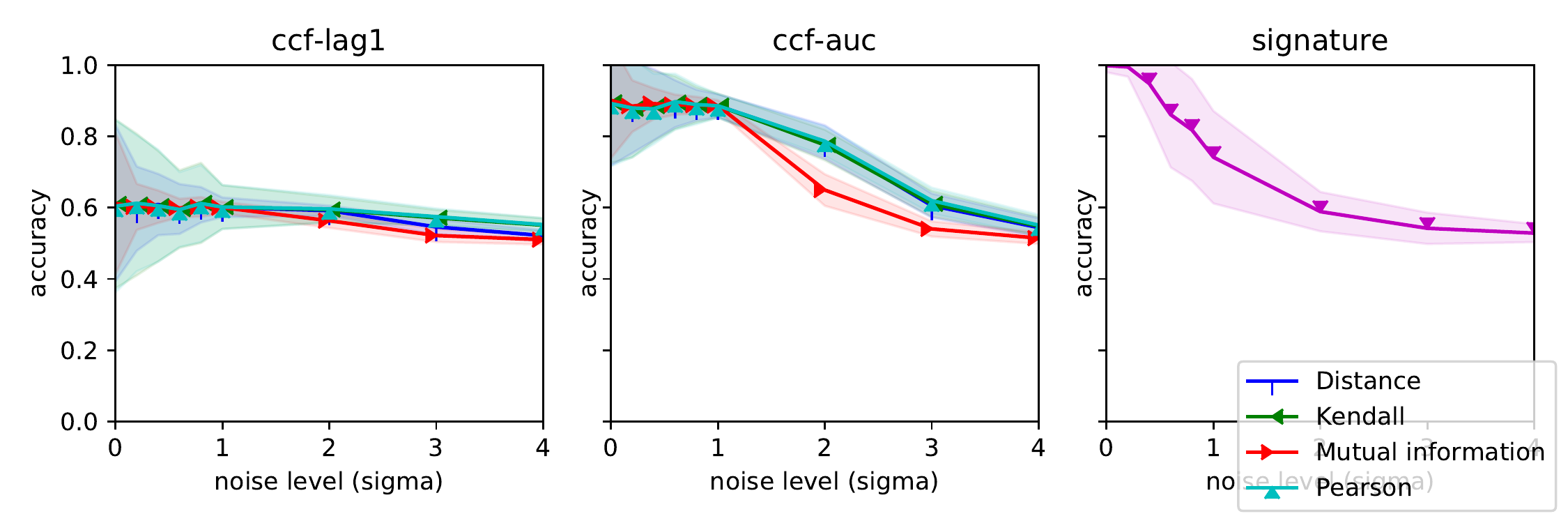}
	\caption{Average and confidence interval for accuracy by lead-lag detection method in the linear setting.}
	\label{fig:synthetic_data_linear_corr}
\end{figure}

\begin{figure}[h!] 
	\centering
		\includegraphics[width=1\textwidth]{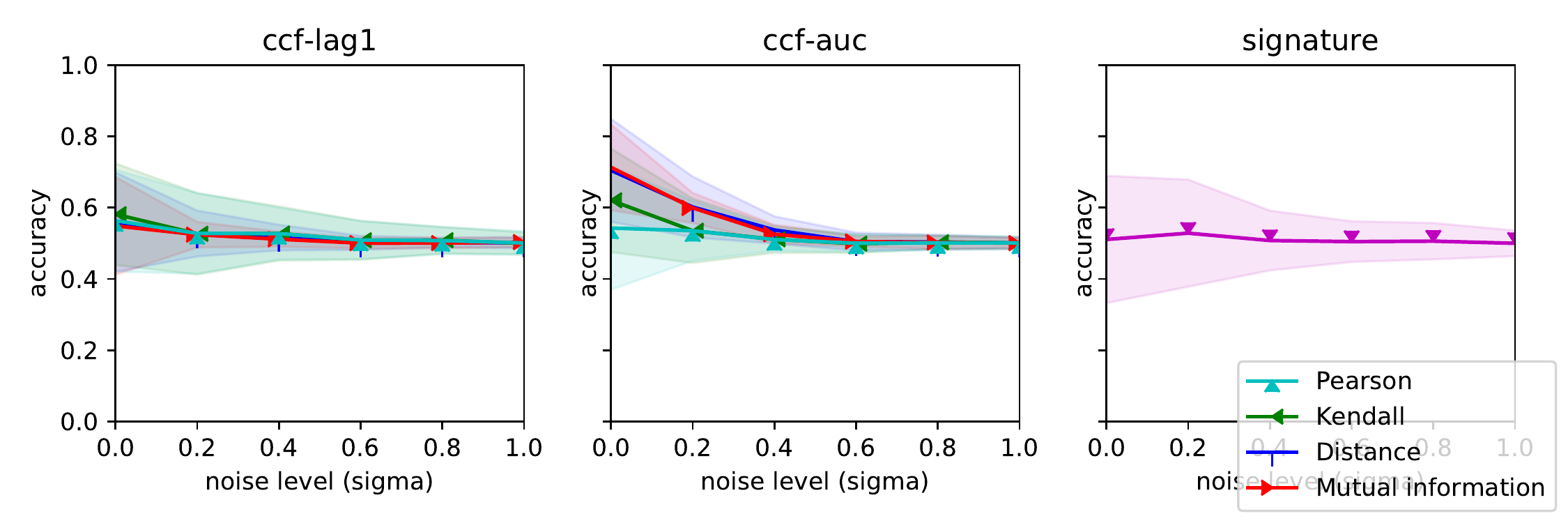}
	\caption{Average and confidence interval for accuracy by lead-lag detection method in the cosine setting.}
	\label{fig:synthetic_data_cosine_corr}
\end{figure}

We display the average value and confidence interval for the lead-lag component detection accuracy over the 48 samples in the linear setting in  \Cref{fig:synthetic_data_linear_corr} and the cosine setting in  \Cref{fig:synthetic_data_cosine_corr}. The confidence interval is a $95\%$ Gaussian for the accuracy computed on a sample from the data-generating process.

Figure \ref{fig:synthetic_data_linear_corr} shows that the proposed lead-lag detection components are able to detect linear lead-lag associations and that their performance decreases to random chance performance as the level of noise in the synthetic data experiment increases. The $\textbf{ccf-auc}$ and signature methods work best in this setting. Within the $\textbf{ccf-auc}$ method, the non-linear Kendall and distance correlation metrics are able to maintain similar performance to the linear metric. The outperformance of the $\textbf{ccf-auc}$ method over the $\textbf{ccf-lag1}$ method shows the advantage of considering a larger number of lags in the cross-correlation function when pairs of time series depend on each other through large lag values.

The performance of the methods decreases in the cosine setting: the noise level at which the performance of all methods drops to that of random chance is about is around $\sigma = 0.5$ (compared with $\sigma = 4$ in the linear setting). In particular, the $\textbf{ccf-lag1}$ and signature methods perform poorly; this is not a surprise since this method cannot deal with negative associations. The $\textbf{ccf-auc}$ method using mutual information or distance correlation is able to achieve the highest accuracy; this illustrates the use of methods that are able to take into account negative and non-linear associations.

\begin{figure}[h!] 
\centering
\includegraphics[width=0.55\textwidth]{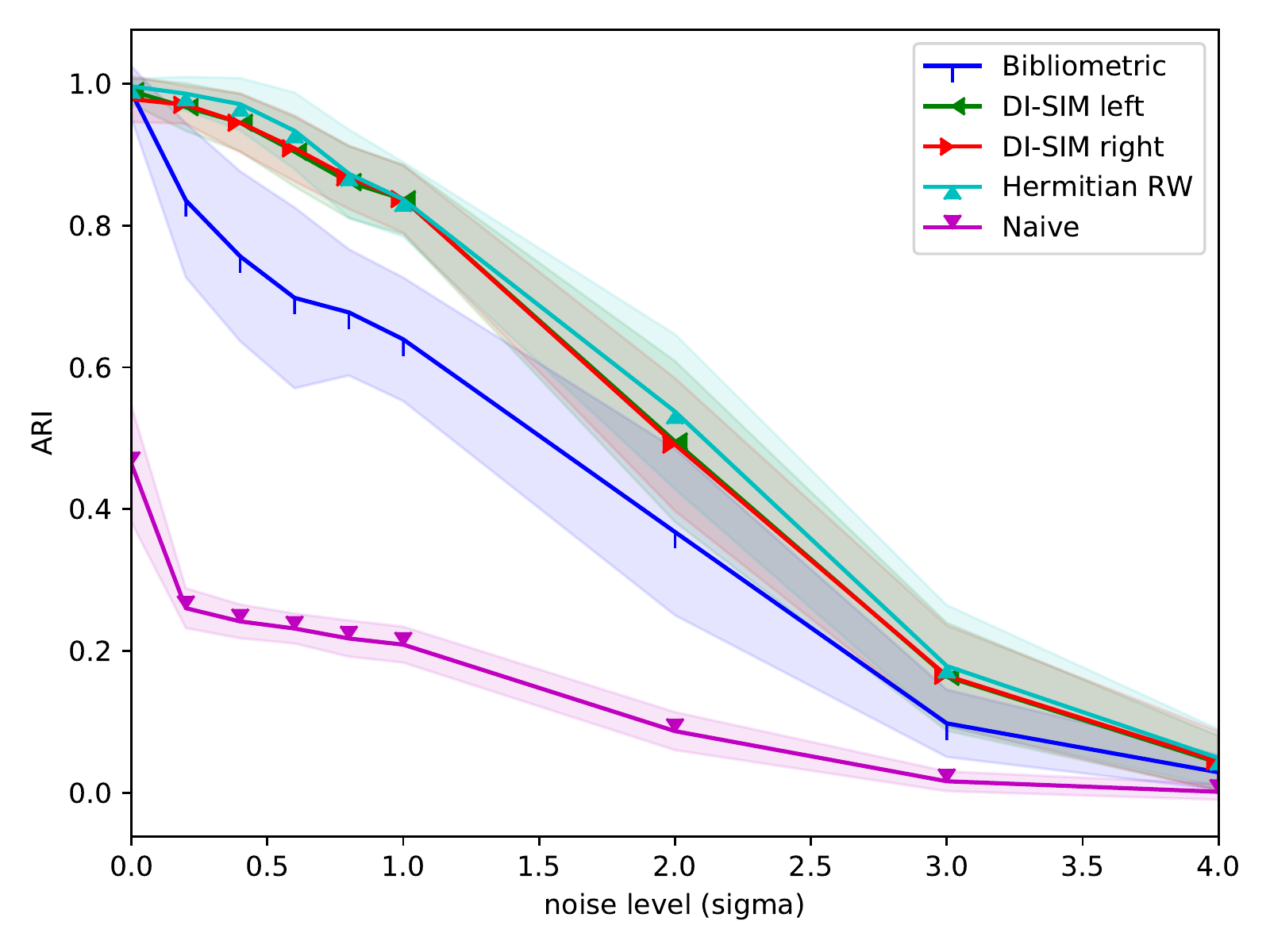}
\caption{Average and confidence interval for the ARI by clustering method in the linear setting.}
\label{fig:synthetic_data_linear}
\end{figure}

\begin{figure}[h!] 
\centering
\includegraphics[width=0.55\textwidth]{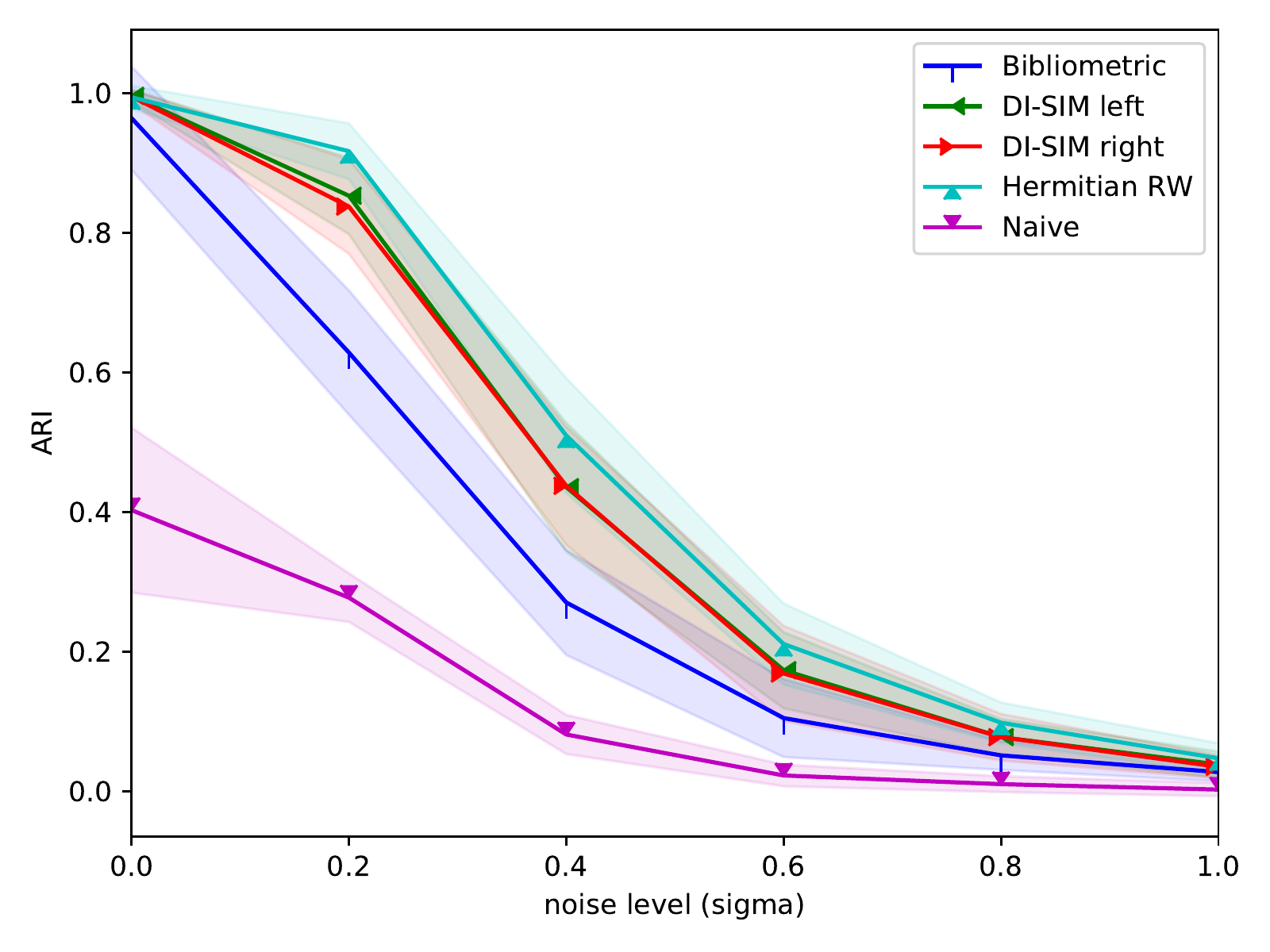}
\caption{Average and confidence interval for the ARI by clustering method in the cosine setting.}
\label{fig:synthetic_data_cosine}
\end{figure}

In order to compare the performance of different clustering methods, we compute, for each clustering method and experimental repetition, the marginal of ARI over the different lead-lag detection metrics. The mean and confidence interval for the ARI values over the experimental repetitions are shown in  \Cref{fig:synthetic_data_linear} and \Cref{fig:synthetic_data_cosine}.   

We observe that the (non-naive) implementations of our method are able to recover almost perfectly (ARI of 1) the clustering in both settings (1) and (2) when $\sigma$ is low. As expected, the performance of our methods decrease as $\sigma$ increases; the performance in the cosine setting decreases faster than in the linear setting. The Hermitian RW and the DI-SIM clustering methods perform best in the settings considered. The Hermitian RW method targets clusters with high imbalance \citep{Cucuringu2019} and is therefore particularly suited to the task of clustering time series according to directed imbalances in their lead-lag relations. The importance of edge direction is illustrated by the relatively poor performance of the naive method, which relies solely on the magnitude and not the direction of the edges. 

Note that even as the number of lags considered in the cross-correlation function by the $\textbf{ccf-lag1}$  and $\textbf{ccf-auc}$  component method (1 and 5 lags, respectively) is lower than the largest lag dependence between any two pairs of time series (e.g. $l_{100} - l_1 = 9$ in the linear setting), our overall two-stage pipeline using these component methods is still able to leverage enough similarities in the dependence structure between the time series to correctly recover the ground-truth clustering. We are able to successfully cluster in this case since ${\max}_{i \in \{1, \ldots, p \}} {\min}_{j \in \{1, \ldots, p \}} \vert l_{i} - l_j\vert = 1$, 
which is less than or equal to the number of lags considered by the cross-correlation function methods.

Our experimental observations are robust to the other synthetic data generating processes reported in \Cref{sec:Appendix}. Similar results are also observed when performing simulation studies for a smaller number of time series and smaller sample sizes.

\subsubsection{Interaction of lead-lag and clustering components}

\begin{figure}[h!] 
	\centering
		\includegraphics[width=1.\textwidth]{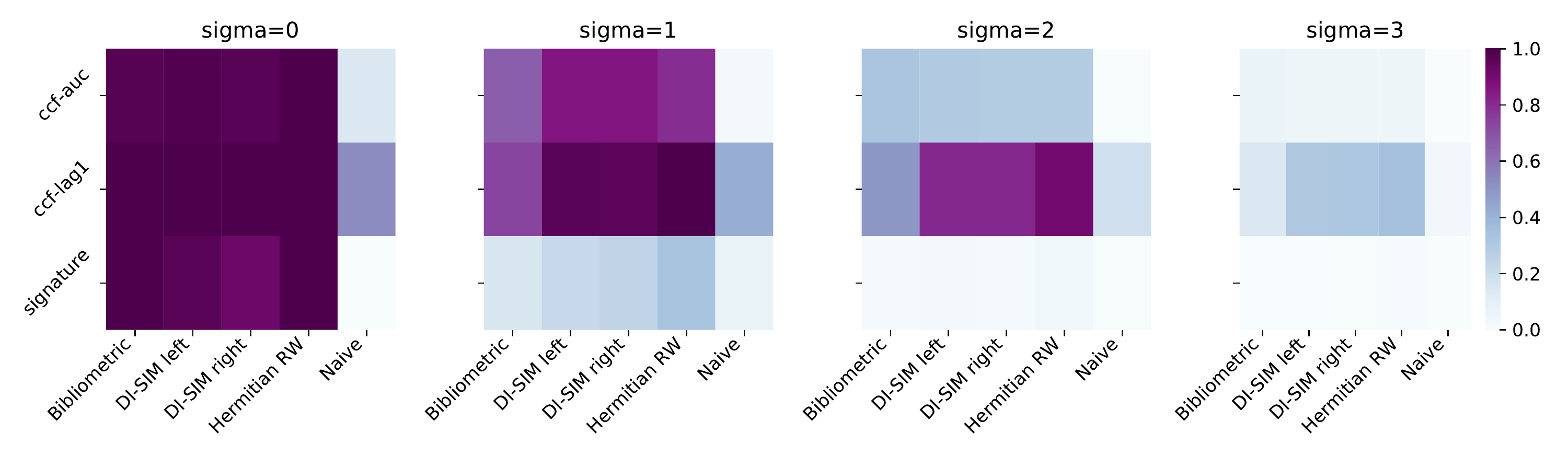}
	\caption{Average ARI by lead-lag and clustering method in the linear setting.}
	\label{fig:interaction_linear}
\end{figure}

In this section, we investigate the joint dependence of the pipeline on the lead-lag and clustering components. The performance of the pipeline, measured by ARI averaged over the different Monte Carlo repetitions, is shown for linear and cosine synthetic data settings in figures  \ref{fig:interaction_linear} and \ref{fig:interaction_cosine}; the other synthetic data settings are presented in Appendix \ref{sec: Appendix - Synthetic data experiment: interaction of lead-lag and clustering components}. For each synthetic data setting, we select a range of noise levels $\sigma$ that are representative of the different levels of overall ARI significance. In these figures, for the pipelines using $\textbf{ccf-lag1}$ and $\textbf{ccf-auc}$ components, we show the ARI averaged across the 4 different choices of sample correlation function described in section \ref{subsubsec:Lead-lag metrics based on a functional of the cross-correlation}.

\begin{figure}[h!] 
	\centering
		\includegraphics[width=1.\textwidth]{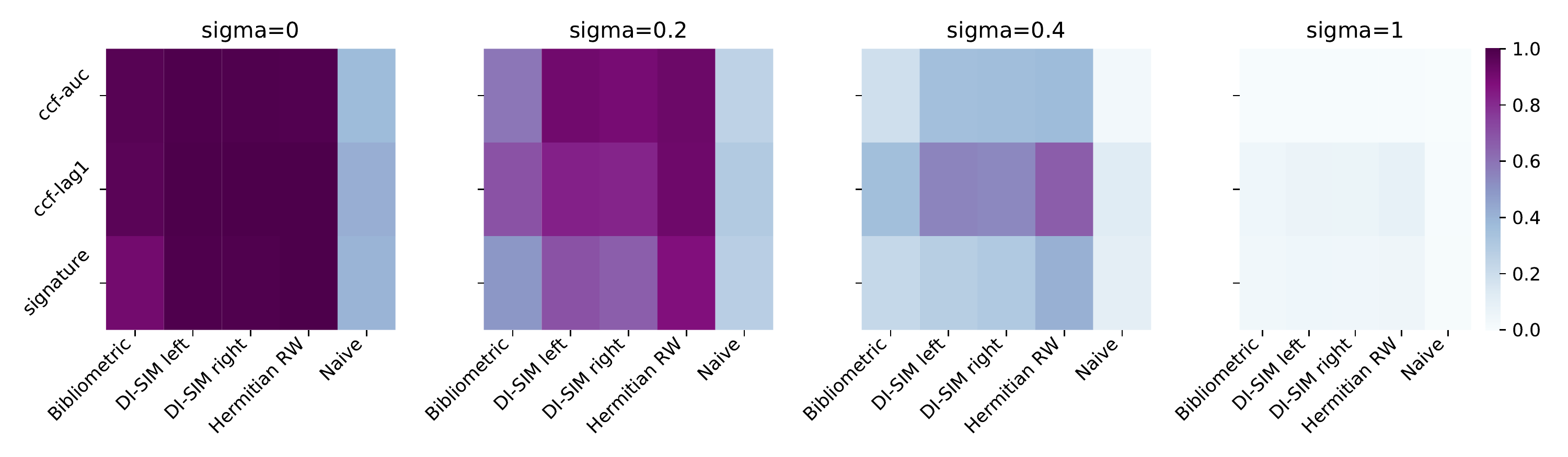}
	\caption{Average ARI by lead-lag extraction and clustering method in the cosine setting.}
	\label{fig:interaction_cosine}
\end{figure}

We observe that the pipelines that use $\textbf{ccf-lag1}$ or $\textbf{ccf-auc}$ lead-lag extraction components with $\textbf{DI-SIM}$ or $\textbf{Hermitian RW}$ as the clustering component tend to perform best. For small values of $\sigma$, the performance of each of these pipelines tends to be quite similar. For larger $\sigma$ values, the relative performance difference between the different lead-lag extraction components tends to increase, with the performance of the DI-SIM and Hermitian RW components within a pipeline using $\textbf{ccf-lag1}$ or $\textbf{ccf-auc}$ appearing to be quite correlated. Eventually, the performance of every pipeline drops to 0 as $\sigma$ increases.

\subsubsection{Ablation study: varying hyperparameter corresponding to the number of clusters}

\begin{figure}[h!] 
	\centering
		\includegraphics[width=0.55\textwidth]{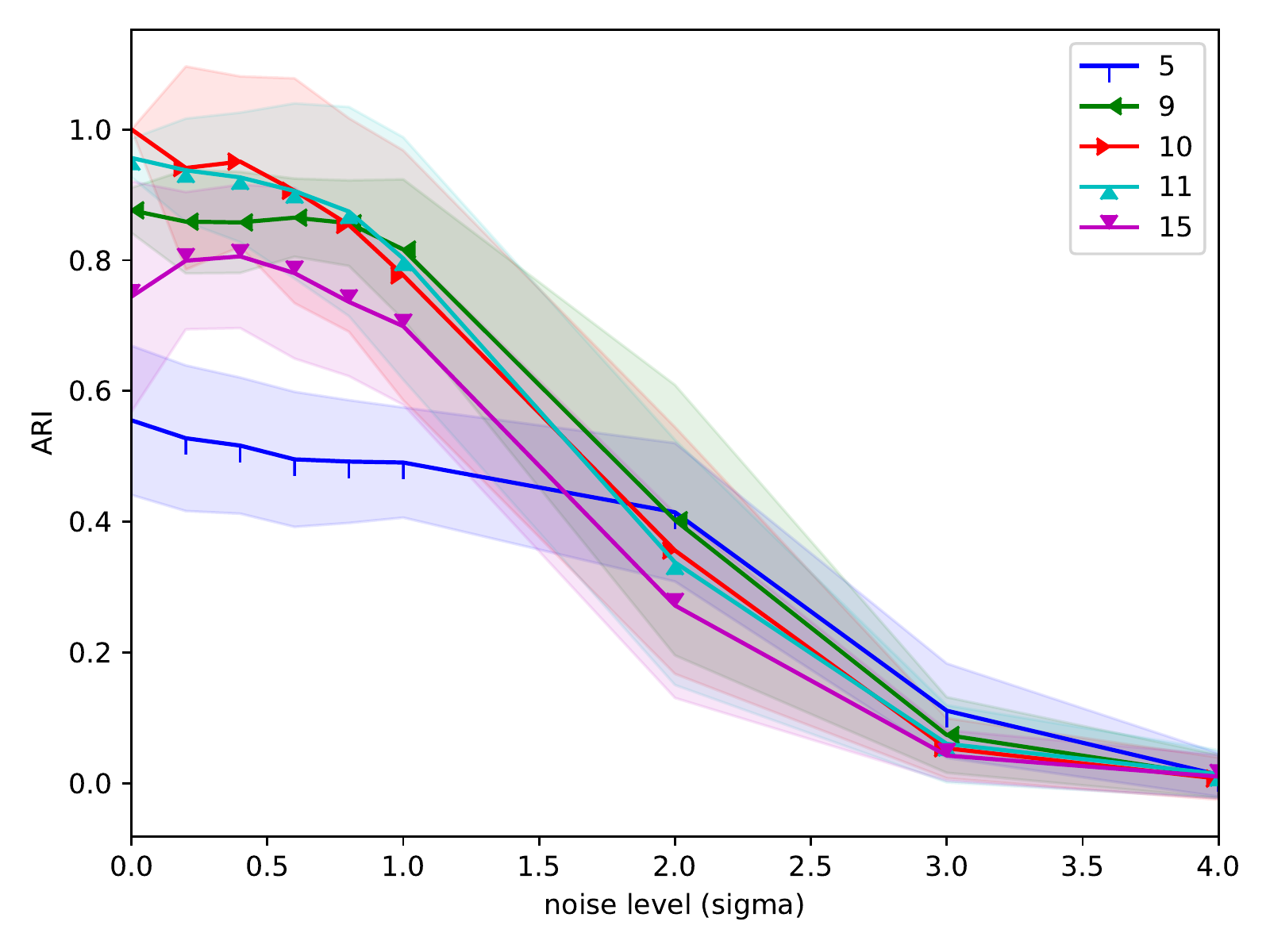}
	\caption{Average and confidence interval for the ARI by different levels of the hyperparameter corresponding to the number of clusters in the linear setting.}
	\label{fig:ablation-linear}
\end{figure}

We perform an ablation study to examine the sensitivity of the pipeline to the hyperparameter controlling the number of clusters returned by the clustering algorithm. In Figure \ref{fig:ablation-linear}, we present the results for the typical linear synthetic data generating setting using a pipeline of $\textbf{ccf-auc}$ with distance correlation and Hermitian RW clustering. Results for the cosine, Legendre and Hermite data settings are shown in the Appendix \ref{sec:Appendix - Synthetic data ablation study: varying hyperparameter corresponding to the number of clusters}.

The true underlying number of clusters in the linear data setting is 10 (see Section \ref{sec:Synthetic data experiment}). In Figure \ref{fig:ablation-linear}, we see that the performance of the pipeline is robust to small variations in the hyperparameter corresponding to the number of clusters around the true underlying number of clusters. Further, we find that using a large hyperparameter value for the number of clusters results in a large decay in the ARI of the pipeline.

\subsubsection{Summary of synthetic data experiment results}

To summarise this section, we have validated our pipeline on five synthetic data generating processes. While the choice of particular correlation components should be driven by the application in mind, the $\textbf{ccf-auc}$ method using distance correlation achieves relatively strong performance both in the linear and in the cosine synthetic data generating settings. The clustering component methods that were found to perform best were the DI-SIM and Hermitian RW methods. 

\section{US equity data experiment} \label{sec:US equity data experiment}

It is well known that US equity returns exhibit a cross-sectional factor structure \citep{Fama1993}. Some of the prominent factors, for example the factors representing industry membership, can exhibit cluster membership. This induces a clustering structure in the synchronous cross-sectional equity returns \citep{Farrell1974}. In addition to this  synchronous clustering structure, we conjecture that there exists a clustering structure in US equities due to inter-temporal relations in equity returns. In this section, our method is applied to construct and cluster a lead-lag network on a US equity universe, and investigate the resulting data-driven clustering. On the basis of a-priori considerations and performance under the synthetic data experiments, a lead-lag metric that computes distance correlation \citep{Szekely2007} between the shifted time series and a directed clustering method that uses the spectrum of a Hermitian adjacency matrix are suitable components for the application of our method to US equity returns. We will use $\textbf{ccf-auc}$ with lags $l \in \{-5, \ldots, 5\}$ with the distance correlation as our lead-lag metric, and Hermitian RW clustering as our clustering step. This method has the potential to capture non-linear lead-lag relations between returns on the scale of up to a week. We set the number of clusters, a hyperparameter of our algorithm, to 10 in order to facilitate comparison with the industry-sector clustering of equities.

\subsection{Data description} \label{sec: US equity data description}
We consider the universe of 5325 NYSE equities spanning from 04-01-2000 to 31-12-2019 from Wharton's CRSP database \citep{WRDS} -- restricting our attention to equities trading on the same exchange to avoid spurious lead-lag effects due to non-synchronous trading \citep{Campbell1997}. The data consists of daily closing prices from which we compute daily log-returns. We also compute the average daily dollar volume that is traded for each equity. We subset to the equities that have the largest average volume (largest 500 equities in average volume) and the least number of missing values (at least 2.5 years' worth of non-missing data). This results in a data set of 434 equities. Filtering to the most traded equities with the least number of missing prices reduces the risk of spurious lead-lag effects due to non-synchronous trading \citep{Campbell1997}. Any remaining missing prices are forward-filled prior to the calculation of log-returns.

\subsection{Data analysis}

\subsubsection{Illustration of US equity lead-lag matrix}

\begin{figure}[h!]
	\centering
		\includegraphics[width=0.6\textwidth]{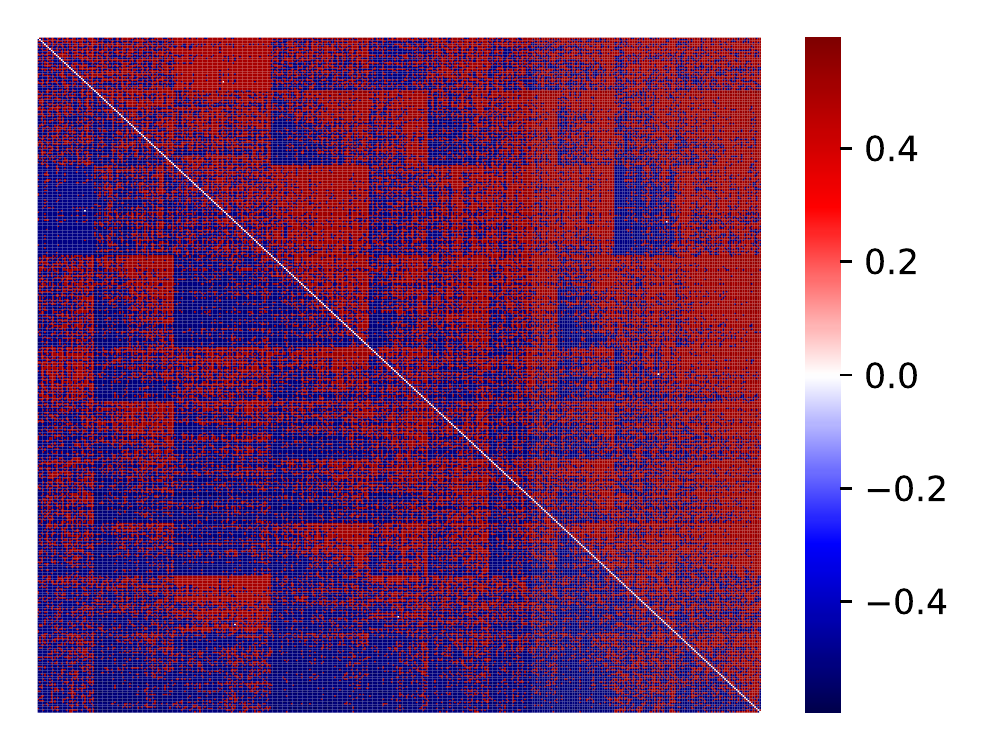}
	\caption{Heatmap of the double-sorted lead-lag $p \times p$ matrix $A - A^T$. The rows and columns of the matrix index the $p=434$ equities, and are categorised by cluster membership (labelled by the leadingness metric \eqref{eqn:cluster_leadingness}). Within each cluster, we sort the equities by their respective row-sum in $A - A^T$, a proxy for their individual leadingness. }   
	\label{fig:heatmap_herm_rw_clustering_ordering}
\end{figure}

\Cref{fig:heatmap_herm_rw_clustering_ordering} shows a sorted skew-symmetric lead-lag matrix encoding the measurement between each pair of stocks.  Positive entries in the matrix correspond to a leading relationship between the stock depicted on the vertical axis with respect to the stock depicted on the horizontal axis. Similarly, negative values indicate that the horizontal axis stock leads the vertical axis one. The skew-symmetric matrix $A - A^T$ depicted in   \Cref{fig:heatmap_herm_rw_clustering_ordering} is double-sorted by the leadingness metric \eqref{eqn:cluster_leadingness} for each cluster and then, within each cluster, by the rowsum $\sum_{j=1}^p\left[ A_{ij} - A_{ji}\right]$ of each equity $i$ that is a member of the cluster. A block structure is apparent, with the last block being a highly lagging cluster.

\subsubsection{Statistical significance testing for lead-lag clusters}

We test whether there is a statistically significant time dependence in daily US equity returns using a permutation test on the spectrum of the Hermitian adjancency matrix $\Tilde{A} = i(A - A^T)$. Under the null hypothesis that there is no time dependence, the ordering of the rows of the daily returns matrix $Y \in \mathbb{R}^{T \times p}$ is drawn uniformly at random from the set of all permutations on $\{1, \ldots, T\}$, $\sigma \in S_T$. Therefore, under the hypothesis of no time dependence, the spectrum of the observed lead-lag matrix should be consistent with the distribution over the spectra of matrices $\{\Tilde{A}_{\sigma} \}_{\sigma \in S_T}$ computed using row-permuted returns matrices $Y_{\sigma(t), j}, t = 1, \ldots, T, \, j = 1, \ldots, p$. Since lead-lag cluster structure is associated with the largest eigenvalues of the Hermitian matrix $\Tilde{A}$ \citep{Cucuringu2019}, our permutation test statistic is set to be the largest eigenvalue of $\Tilde{A}$. We use 200 Monte Carlo samples from the null distribution. Under the null hypothesis, the Monte Carlo probability that the largest eigenvalue is greater than or equal to the observed largest eigenvalue is $1/201$. We thus reject the null hypothesis with p-value $p < 0.005$, and conclude that there is significant temporal structure in US equity markets.  

Note that a rejection of the null implies either
\begin{enumerate}
    \item Significant auto-correlation
    \item Significant cross-correlation
    \item Some combination of 1. and 2.
\end{enumerate}

It is not possible to resolve the identification issue between these three cases using our method. However, since our test statistic is a summary statistic of the lead-lag matrix spectrum, which encodes cross-correlations between time series and relates to the clustering structure \citep{Cucuringu2019}, a rejection of the null \textit{suggests} that there is significant cluster structure in the lead-lag matrix. Our statistically significant results when using our method for downstream prediction tasks (which relies solely on cross-equity prediction and not auto-correlation) in  \Cref{sec:Financial forecasting application} provide further evidence for significant clustered lead-lag structure in the US equities.

\subsubsection{Comparing data-driven clustering with known lead-lag mechanisms} \label{sec:Comparing data-driven clustering with known lead-lag mechanisms}

We investigate whether our data-driven lead-lag extraction and clustering results can be explained by three potential mechanisms in the empirical finance lead-lag literature.

\begin{enumerate}
    \item Sector membership induces clustered lead-lag effects. \cite{Biely2008} find associations between sector membership and lead-lag structure on the high-frequency scale of returns.
    \item Equities with higher trading volume are hypothesised to lead lower volume equities. The disparities in trading volume across equities can lead to non-synchronous trading 
    lead-lag effects \citep{Chordia2000, Campbell1997}. Clustering structure may be induced by ordering equities based on quantiles of average trading volume. 
    \item Larger capitalisation equities are hypothesised to lead lower capitalisation equities
    \citep{Lo1990}. This market capitalisation mechanism can produce lead-lag effects partly via non-trading effects and partly via other channels \citep{Campbell1997}. \cite{Conrad1991} also find that large stocks may lead small stocks via volatility spillovers. Clustering structure may be induced by ordering equities based on quantiles of market capitalisation.
\end{enumerate}

\paragraph{Comparison of data-driven clustering with industry membership clustering}

We compute the Jaccard similarity coefficient between the data-driven Hermitian RW clustering and the clustering due to industry membership. We use the first level of the Standard Industrial Classification (SIC) \citep{WRDS}
code for the firm corresponding to each equity in order to assign the equity to an industry. Table \ref{tab:sic_count} counts the number of equities that are a member of each SIC sector. Most sectors have a relatively large number of equities, with {\it Agriculture, Forestry and Fisheries} and {\it Services} being quite small. 

\begin{table}  [h!]
\centering
\begin{tabular}{ll}
\hline
Retail                  & 90 \\
Manufacturing           & 67 \\
Construction            & 66 \\
Mining                  & 58 \\
Trans., Util. \& other  & 54 \\
Fin., Ins. \& RE        & 46 \\
Wholesale               & 43 \\
Services                & 9  \\
Agri., Forest. \& Fish. & 1 \\
\hline
\end{tabular}
\caption{Number of equities in each SIC industry sector.}
\label{tab:sic_count}
\end{table}

\begin{table}[h!]
\centering
\begin{tabular}{ccc c c c c c cc}
\hline
0& 1 & 2 & 3 & 4 & 5& 6& 7& 8& 9\\
37    & 49    & 57  & 58   & 35   & 35    & 42  & 34 & 32  & 55 \\
\hline
& \\
\end{tabular}
\caption{Number of equities in each Hermitian RW cluster. Cluster ID is shown in the top row.}
\label{tab:hermitian_count}
\end{table}

For comparison, the number of equities in each of the Hermitian RW clusters is shown in Table \ref{tab:hermitian_count}. The Hermitian RW algorithm leads to clusters of approximately equal size. 

\begin{figure}[h!] 
\centering
\includegraphics[width=0.65\textwidth]{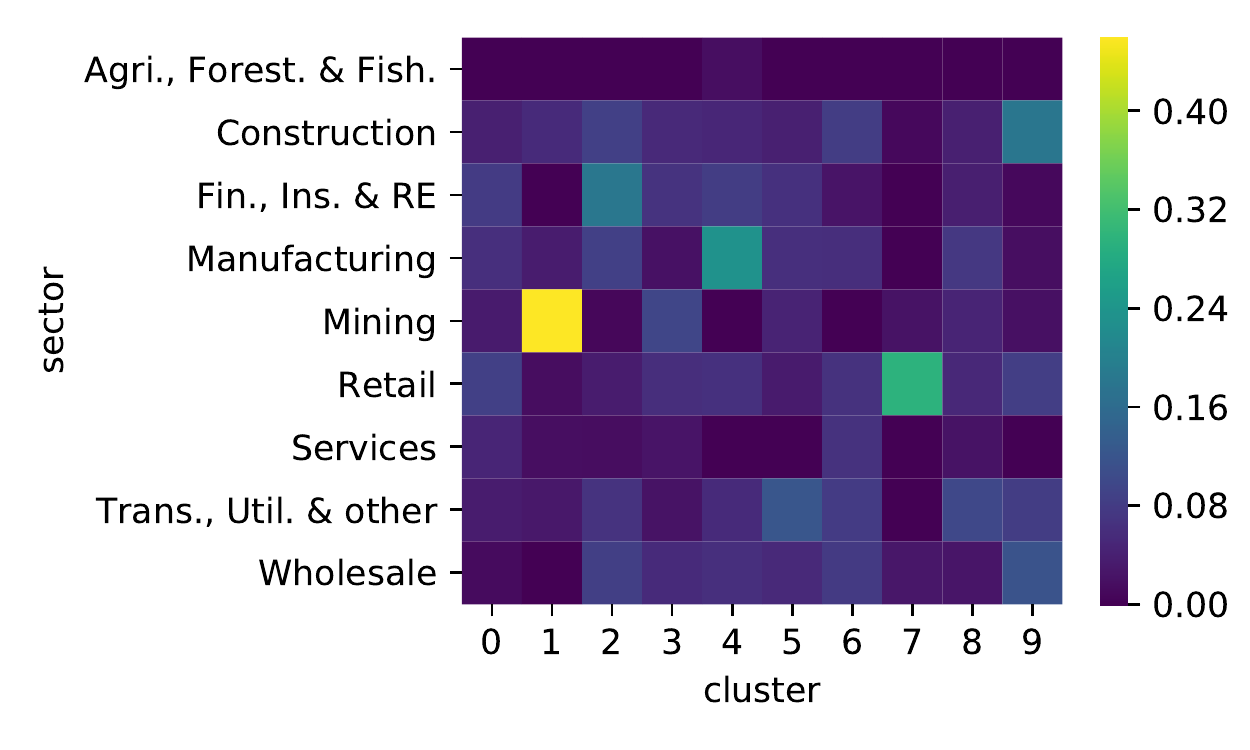}
\caption{The Jaccard similarity coefficient between the Hermitian RW  clusters and industry clusters (SIC).}
\label{fig:jaccard_index_sectors_clusters}
\end{figure}

Figure \ref{fig:jaccard_index_sectors_clusters} displays the Jaccard similarity between each pair of Hermitian RW and industry clusters. Overall, given the low values of the Jaccard similarity coefficients, the clustering seems to recover a structure that goes beyond simple industry sectors. 

\begin{figure}[h!] 
\centering  
\begin{minipage}{.48\textwidth}
\centering  
\includegraphics[width=0.75\textwidth,trim=0.5cm 0.5cm 0.5cm 0.5cm,clip]{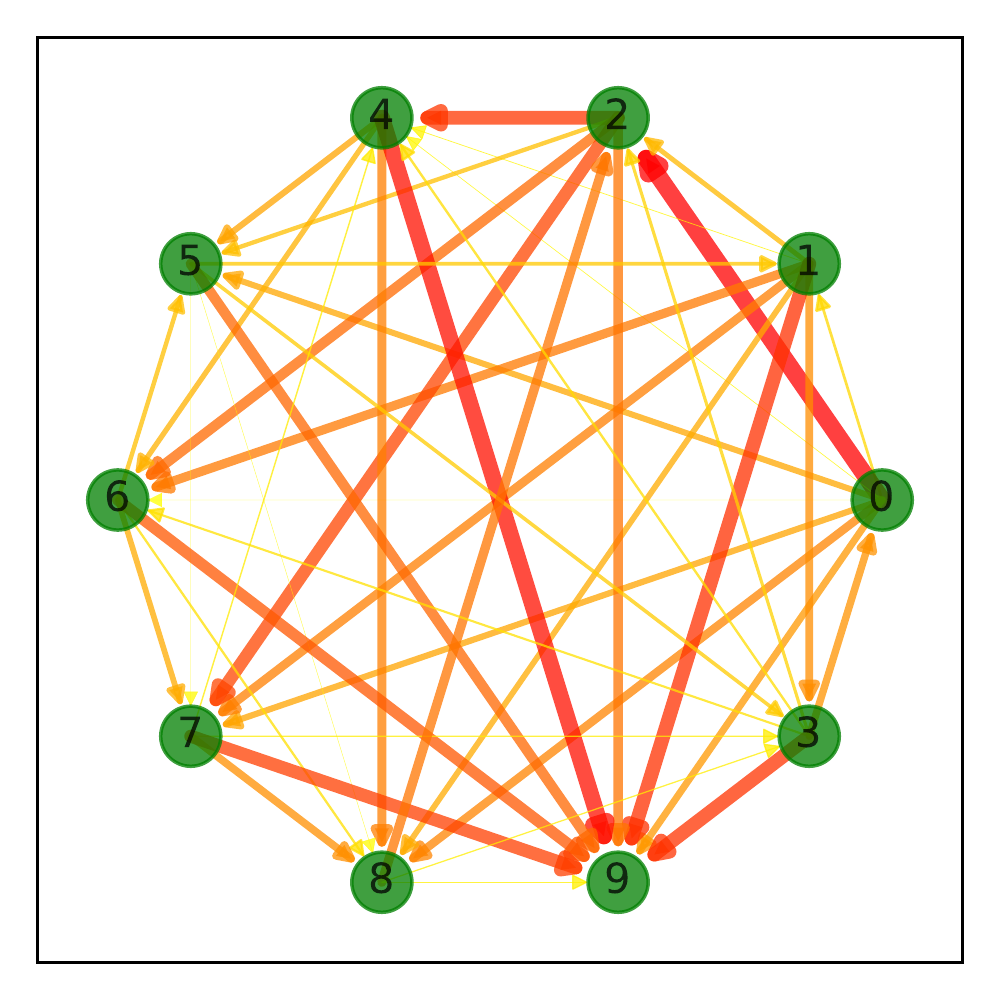}
\caption{Meta-flow network for Hermitian RW clusters; clusters are represented by nodes and larger edge weights are depicted by bolder colours and thicker lines.}
\label{fig:meta-flow_network}
\end{minipage}
\hspace{4mm}
\begin{minipage}{.48\textwidth}
	\centering
		\includegraphics[width=0.99\textwidth]{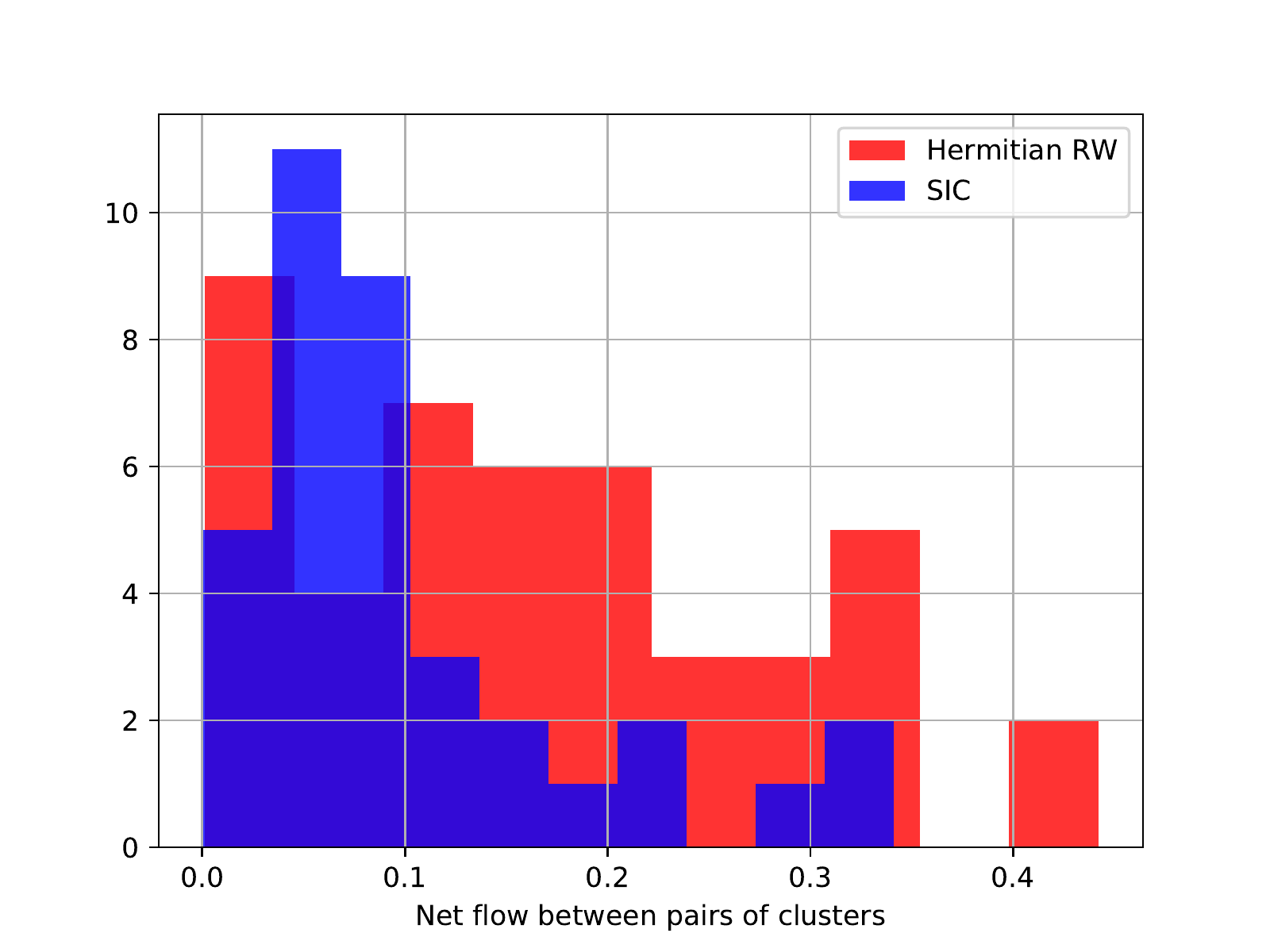}
	\caption{Histogram of Hermitian RW and SIC clustering meta-flow network edge weights. The edge colours are layered in a semi-transparent fashion.
	}
	\label{fig:hermitian-rw_sic_histogram}
\end{minipage}
\end{figure}

However, there does appear to be some association between certain SIC sectors and Hermitian RW clusters. 
We observe that the Mining sector seems to be strongly associated with cluster 1 (the second most leading cluster). The Finance, Insurance and Real Estate sector is also associated with a relatively leading cluster (cluster 2). These observations are consistent with the findings of \cite{Biely2008} that the finance and energy sectors have strong participation in the significant eigenvalues of the lead-lag matrix\footnote{While \cite{Biely2008} use GICS sector classification in their analysis, the GICS Energy sector has substantial overlap with the Mining SIC sector.}. \cite{Xia2018} also find that the Financial and Real Estate sectors are associated with leading equities in the Chinese equity market. These associations between SIC code and Hermitian RW membership provide a partial interpretation for the links of the meta-flow network corresponding to the Hermitian RW clustering. The meta-flow network is depicted in   \Cref{fig:meta-flow_network}. For example, we see that one of the strongest flows is from cluster 4 to 9 -- which are associated with Manufacturing and Construction respectively.

Figure \ref{fig:hermitian-rw_sic_histogram} displays a histogram of the edge weights of
two meta-flow networks: one corresponding to Hermitian RW clustering and the other corresponding to SIC clustering. The data-driven Hermitian RW clustering results in larger flow between pairs of clusters than an industry-based clustering. This demonstrates the efficacy of our method in retrieving pairs of clusters with high flow imbalance.

\paragraph{Comparing data-driven clustering with market capitalisation and volume-based explanations} \label{paragraph:Comparing data-driven clustering with market capitalisation and volume-based explanations}

\begin{figure}[h!]
\centering
\begin{minipage}{.45\textwidth}
\centering  
\includegraphics[width=0.95\textwidth]{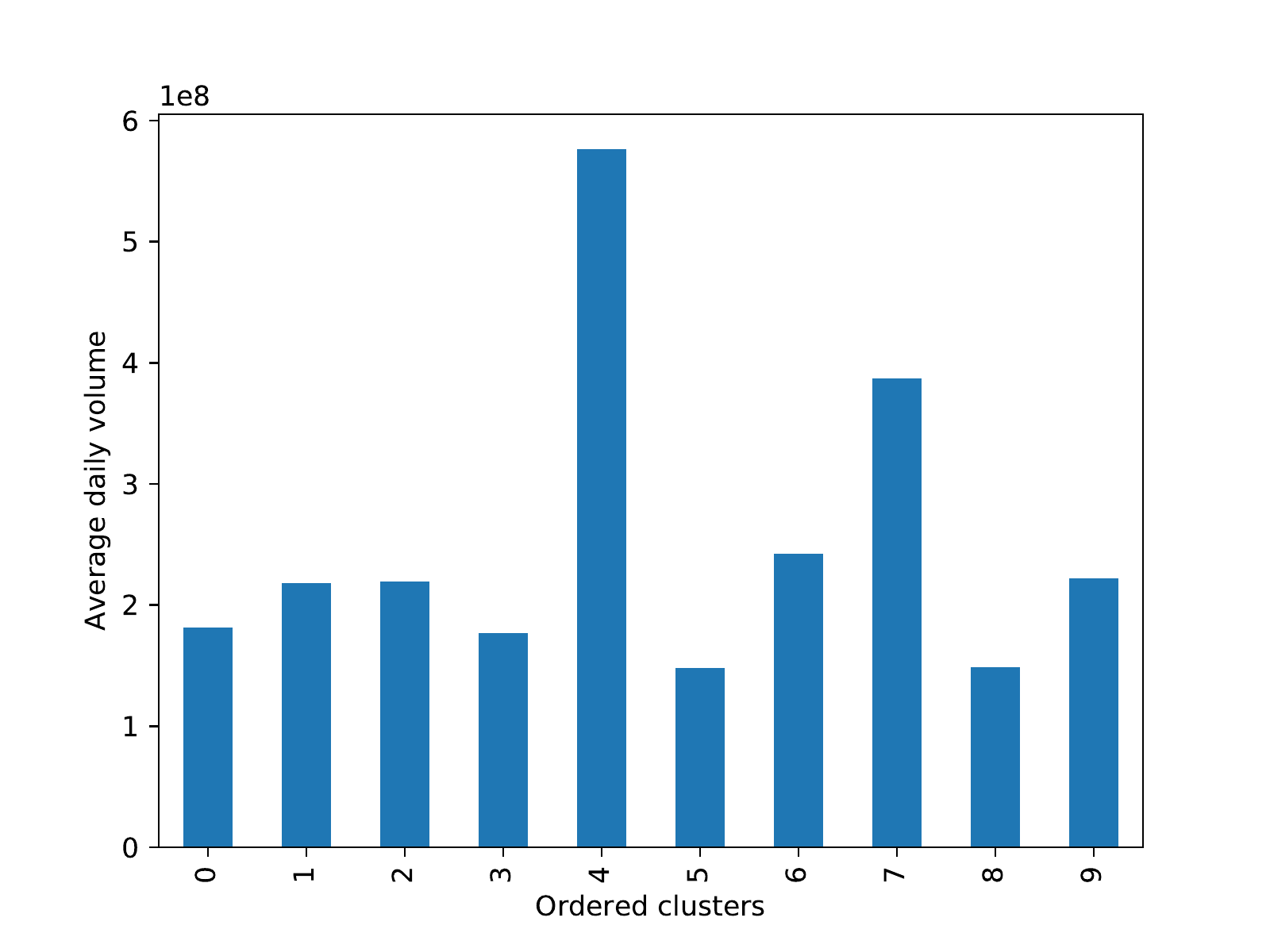}
\caption{Average daily dollar volume by Hermitian RW cluster.}
\label{fig:bar_plot_average_daily_volume}
\end{minipage}
\hspace{7mm}
\begin{minipage}{.45\textwidth} 
\centering
\includegraphics[width=0.95\textwidth]{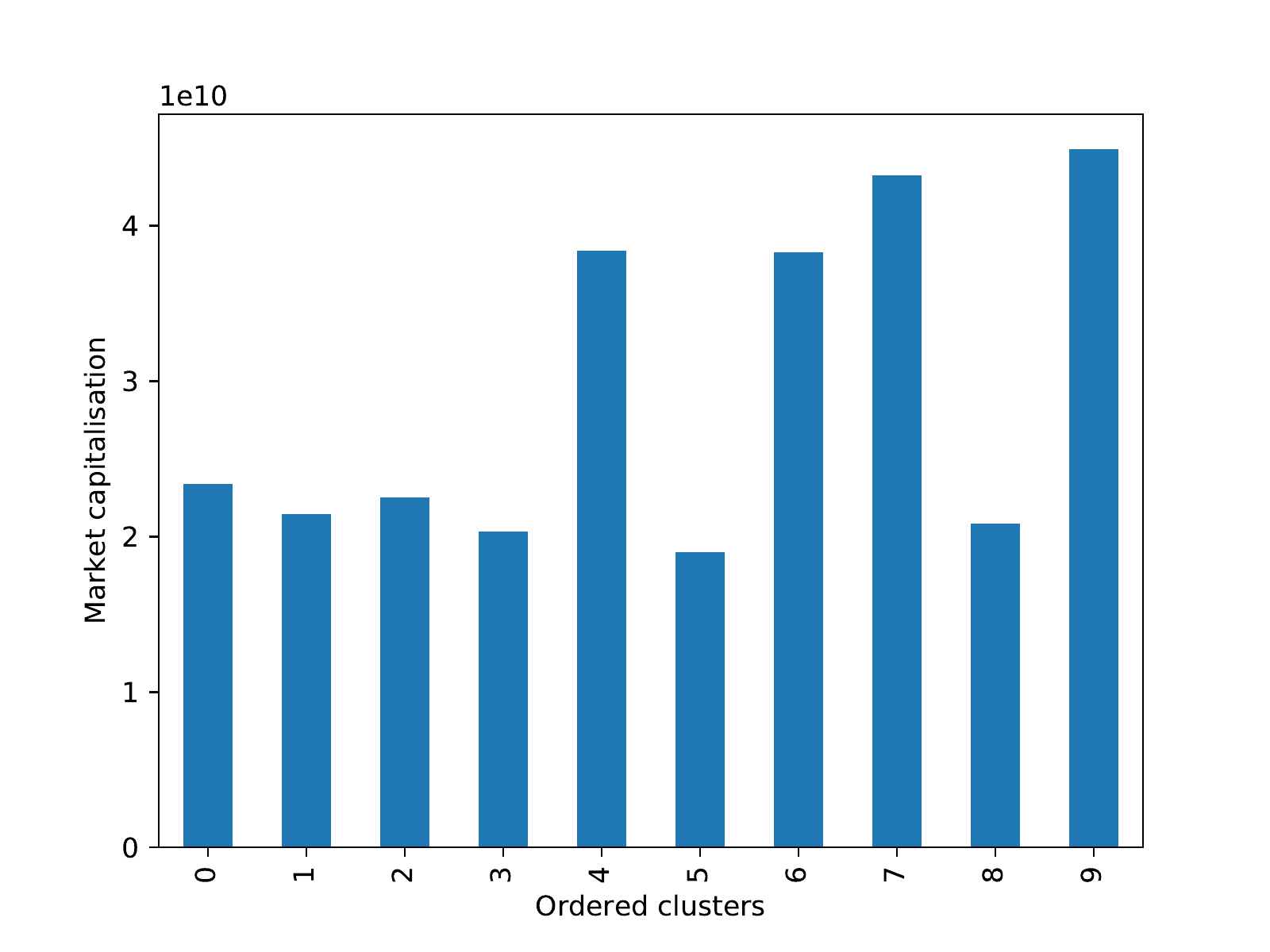}
\caption{Average market capitalisation by Hermitian RW cluster.}
\label{fig:bar_plot_market_capitalisation}
\end{minipage}
\end{figure}

Figures \ref{fig:bar_plot_average_daily_volume} and \ref{fig:bar_plot_market_capitalisation} display the average daily dollar volume and market capitalisation averaged across all stocks in a given cluster. We observe that the leading clusters (clusters labelled $0-3$) do not appear to have larger average daily dollar volume or market capitalisation.

In order to examine  the association between the tendency for an equity to lead and its daily dollar volume or market capitalisation at a sub-cluster level, we compute the Spearman correlation between the row-sums of the lead-lag matrix -- which provides a metric for the tendency of each cluster to lead -- and these equity characteristics (trading volume and market capitalisation). This results in a Spearman correlation of 0.01 and -0.15 between the lead-lag row-sums and the equity trading volume and market capitalisation, respectively. 
These results are not consistent with a positive association between a cluster's tendency to lead and the trading volume or market capitalisation of its constituents. 

Therefore, the results obtained by our data-driven clustering method cannot be explained by the three previously hypothesised mechanisms outlined in   \Cref{sec:Comparing data-driven clustering with known lead-lag mechanisms}. Our proposed method may prove to be useful in the exploration of novel lead-lag mechanisms in the empirical finance community.

\subsection{Time-variation in clusters} \label{sec:Time-variation in clusters}

\begin{figure}[h!]
	\centering
		\includegraphics[width=0.6\textwidth]{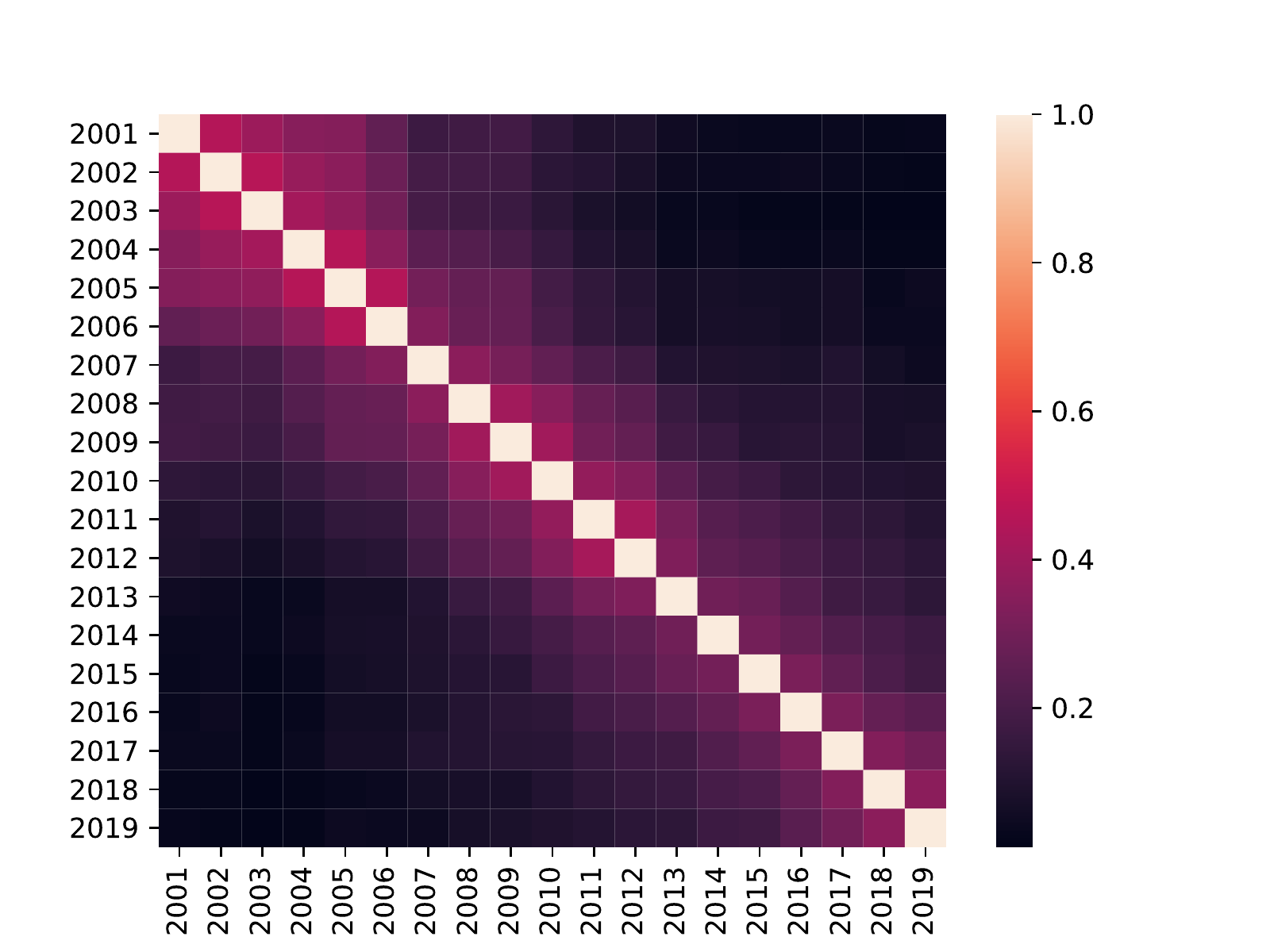}
	\caption{Adjusted Rand index between clusters computed on yearly snapshots of data.}
	\label{fig:yearly_ari}
\end{figure}

To investigate the time-variation in the clustering obtained from our method, we recompute the clustering year-by-year using only data from the retrospective year to do so. In order to compare the similarity in clusterings across time, we calculate the Adjusted Rand Index (ARI) between each pair of yearly clusterings. The results are illustrated in  \Cref{fig:yearly_ari}.

The relatively low ARI values between pairs of clusters indicates some -- albeit low -- persistence in year-to-year lead-lag structure. \cite{Biely2008} find that there is significant persistence in lead-lag structures across time. \cite{Xia2018} agree with our observations and find that the lead-lag phenomenon between two stocks is not constant but emerges during certain periods. They find that, on average, individual lead-lag relationships tend to last for around a year.
 
Further, we see that higher ARI values occur in earlier years: this suggests that there is a decrease in persistence between clusterings as time increases. Nevertheless, in   \Cref{sec:Financial forecasting application}, we show that there is sufficient persistence in the lead-lag cluster relationships in order for a dynamically updated clustering to be useful for forecasting purposes on a daily scale.

\subsection{Limitations and implications of the empirical analysis}

Our novel lead-lag extraction and clustering method yields clusters that cannot be explained by three previously considered mechanisms for lead-lag structure in US equity markets. Below, we discuss the limitations of our empirical analysis and its implications for understanding lead-lag structure in US equity markets.

First, a caveat of our empirical analysis is the instability of the lead-lag structure across time. In   \Cref{sec:Time-variation in clusters}, we observe that the lead-lag structure does not exhibit high overall persistence. Since the lead-lag structure is not stable year-to-year, it is possible that the lead-lag results can be partially explained by the three  mechanisms on a subset of the data. However, we have repeated our empirical clustering analysis on the relatively stable range\footnote{Cf relatively large values of ARI displayed during 2000-2006 in  \Cref{fig:yearly_ari}.} 2000-2006 and have found that the industry-based clustering is unable to fully explain the resulting data-driven clustering on this subset of the data. Furthermore, we have repeated the Spearman correlation analysis that was described in Section \ref{paragraph:Comparing data-driven clustering with market capitalisation and volume-based explanations} using yearly snapshots of data. Appendix Figures \ref{fig:rolling_rowsum_average_daily_volume_corr} and \ref{fig:rolling_rowsum_market_capitalisation_corr} display the Spearman correlation between an equity's tendency to be a leader (which is given by its lead-lag matrix row-sum) and its market capitalisation or trading volume. As these   figures suggest, the association between an equity's tendency to be a leader and its market capitalisation or trading volume is not stable throughout time. There appear to be some periods when the sign of the association is consistent with the positive association predicted by the trading volume and market capitalisation lead-lag mechanisms. Nevertheless, the general sign and transience of the association across time does not support trading volume and market capitalisation as mechanisms which can explain the observed lead-lag structure.
 
A second caveat for the interpretation of our results concerns the relevancy of the market capitalisation mechanism. As explained in   \Cref{sec: US equity data description}, we have restricted our attention to large capitalisation equities in order to avoid non-synchronous trading effects. Therefore, any interpretation of our empirical results must be conditioned by the large capitalisation of our equity universe. In particular, the market capitalisation mechanism may not be relevant under the condition that we restrict attention to the largest equities. In addition, previous papers that have found that smaller cap equities are able to lead larger cap equities if these smaller cap equities receive more news coverage \citep{Scherbina2013}. Thus, the hypothesised market capitalisation mechanism can be modulated by other information diffusion channels. This implies
that the market capitalisation mechanism does not necessarily manifest itself in a positive association between market capitalisation and the tendency of an equity to be a leader.

Thirdly, the lead-lag literature contains other mechanisms that could potentially explain our results \citep{Badrinath1995, Brennan1993, Menzly2010,Cohen2008}. For example, cross-firm information flows through supplier networks have been hypothesised as lead-lag mechanisms \citep{Menzly2010,Cohen2008}.  Testing these and other hypothesised mechanisms as sources for our observed lead-lag results remains further work.

Finally, given the novelty of our method and the fact that the resulting lead-lag structure cannot be explained through the three hypotheses that we have tested, our method may prove to be useful in the exploration of new mechanisms. The use of non-linear lead-lag metrics and effective algorithms for clustering directed networks (such as the distance correlation lead-lag metric and Hermitian RW algorithm) may illuminate lead-lag structures in US equity markets that cannot be explained by existing lead-lag mechanisms in the empirical finance literature.

\section{Financial forecasting application} \label{sec:Financial forecasting application}

Our method of identifying and clustering lead-lag correlation networks can be used for variable selection and feature extraction to facilitate downstream predictive modelling tasks in high-dimensional time series systems. One of the main difficulties in the predictive modelling of high-dimensional systems is variable selection. On the one hand, the selection of too few conditioning variables can result in poor predictive power due to not capturing temporal dependence between the response variable and relevant omitted variables. On the other hand, conditioning on too many variables can lead to the inclusion of many irrelevant variables; this dilutes the predictive power of the model \citep{Runge2017}.

In general, our unsupervised learning method can be used as a variable selection or feature extraction step to inform the choice of potential target and feature variables in a predictive model. This is achieved by using clusters with large net inflows (lagging clusters) to guide the selection of \textit{target variables}, and clusters with large net outflows (leading clusters) guide the selection of \textit{feature variables}. For example, in the latent variable synthetic data generating model presented in Section \ref{sec:Synthetic data experiment}, the method identifies clusters of variables sharing the same lagged dependence on the latent variable $z$. By averaging the time series variables within each cluster,  $\frac{1}{\vert C_i \vert} \sum_{j \in C_i} y^j_t, \, \forall i \in \{1, \ldots, k\}$, the leading latent function $g_1(z_t)$ at time $t$ (the average of time series values in the most leading cluster) and the lagged latent functions $g_l(z_{t-l}), \, l=2, \ldots, k$ (the average of time series values in lagging clusters) can be recovered for each $t = 1, \ldots, T$ thanks to the reduction in observation noise resulting from the averaging procedure. By fitting models that capture the relations between the average value of the lagging clusters (target variable) and the average value of the leading cluster (feature variable), the latent variable dynamics can be captured, allowing the user to make predictions on the subsequent values of the lagging clusters.

When a downstream model is built to capture the relationships between such target and feature variables, it is likely to exhibit stronger predictive power since our method has screened potential explanatory variables. Our method identifies predictable response variables and diminishes the risk of conditioning on irrelevant variables when used in downstream predictive modelling in high-dimensional time series systems. This variable selection approach is useful for the application of returns forecasting in the US equity universe, since this is a highly noisy multivariate time series system where statistical lead-lag effect sizes are weak\footnote{due to the Efficient Markets Hypothesis \citep{Malkiel1970}.}.

We assess the predictive power of our lead-lag extraction and spectral clustering approach by evaluating the out-of-sample performance of a trading signal that was constructed using our method. The risk-adjusted returns of our trading signal will be evaluated using the Sharpe Ratio. In order to test whether the signal's Sharpe Ratio is significantly different to 0, we use a hypothesis test \citep{Opdyke2007} that holds asymptotically under the general conditions of stationary and ergodic signal returns.

Our approach to quantifying the predictive performance of our method by studying the risk-adjusted performance of a portfolio constructed using our method is common in the quantitative finance literature \citep{Asness2013}. The task of constructing a statistically significant trading signal using only publicly available price data in a highly liquid market such as the US equity market is a challenging task due to the informational efficiency of such markets \citep{Malkiel1970}. The weak-form of the Efficient Markets Hypothesis states \citep{Malkiel1970} that markets fully reflect all historical price data; this implies that it is not possible to make economic profits in excess of market equilibrium profits by trading on the basis of such historical price data. The number of empirical studies \citep{Malkiel1970} in strong support of the weak-form of the Efficient Markets Hypothesis underlines the informational efficiency of US equity markets and hence the challenge of constructing a statistically significantly profitable trading signal.

Similarly, \cite{Curme2015a} argue for the use of lead-lag networks to guide variable selection for downstream financial forecasting tasks. However, our results are stronger as we test the performance of our lead-lag network method for variable subset selection in a rolling out-of-sample evaluation.

\subsection{Signal construction} \label{Signal construction}

We keep the trading signal relatively simple in order to effectively assess the predictive performance of the underlying signal derived from our lead-lag extraction and clustering methodology. Our trading signal forecasts lagging cluster returns using smoothed leading cluster returns. In order to evaluate the out-of-sample performance of our method, we compute the clustering $C_1, \ldots, C_k$ and flow graph $F$ on a rolling basis using a 2-month update period and yearly look-back window. Further, using the same update frequency and yearly look-back window, we fit a separate linear model for each pair of clusters. In particular, for each ordered pair of clusters $i, j \in \{1, \ldots, k\}$, we fit a linear model to forecast the mean daily return for lagging cluster $j$
$$
    y_t^{(j)} = \frac{1}{\vert C_j \vert}\sum_{n \in C_j} y_t^{(n)}, 
$$
using an exponentially weighted moving average of the mean returns for cluster $i$ as the covariate (input variable to the linear regression)
$$
    x_t^{(i)} = \frac{1}{\vert C_i\vert }\sum_{n \in C_i} \sum_{l = 1}^{t} \left( 1 - \alpha \right) ^ {l - 1} y_{t - l}^{(n)}.
$$
The choice of exponential parameter $\alpha = 0.4$ assigns $92\%$ of the total weight of the exponential sum $\sum_{l = 1}^{\infty} \left( 1 - \alpha \right) ^ {l - 1}$ to the first 5 lags $l = 1, \ldots, 5$. Thus, the exponential moving average mainly captures lead-lag effects on the scale of approximately up to 1 week, while emphasising higher-frequency daily lead-lag effects. The coefficient $\theta_{ij}$ of the linear model $y^{(j)} = \theta_{ij} x_t^{i}$ is fitted using ordinary least squares\footnote{Note that unbiasedness and consistency do not hold in general for this ordinary least squares estimation due to network effects within the residual error structure.}.

For every day $t = 1, \ldots, T$, we compute the predictive signal from cluster $i$ to $j$ for each ordered pair $i, j \in \{ 1, \ldots, k\}$ of clusters
$$
    \hat{y}_t^{(j)} = \theta_{ij} x_t^{(i)}.
$$
These predictive signals are aggregated using a thresholded flow graph $\Tilde{F}$ where $\Tilde{F}_{ij} = \mathbbm{1} \left\{ F_{ij} > c\right\}$ where $c$ is the $90\%$ quantile of the edge weights of the flow graph $F$. Thus, the flow graph ensures that only the cluster-to-cluster relationships that have shown the greatest historical flow are included in the construction of the signal. Mathematically, the predictive signal $S_t$ for cluster $j \in \{ 1, \ldots, k \}$ is given by 
$$
    S_t(j) = \mathrm{sign}\left( \sum_{n = 1}^k \Tilde{F}_{ij} \hat{y}_t^{(j)} \right).
$$
The signal for a specific equity $m \in \{1, \ldots, p\}$ on day $t$ is set to be the signal for its cluster $C_m$ i.e. $S_t(C_m)$. 

Finally, the signals for each equity are normalised by a 21-day historical rolling estimator of the overall signal's volatility. This rolling normalisation ensures that the overall position size is dynamically adjusted to target a constant 10\% annual volatility. Assuming that the Sharpe Ratio of our signal is constant throughout time, this procedure can be seen as targeting an optimal Kelly criterion \citep{Thorp2011} for the signal on a rolling basis. Further, volatility normalisation tends to bring our daily trading returns closer to stationarity while decreasing their absolute skew and kurtosis; this makes the analysis of our trading returns more reliable.

\subsection{Results}

\begin{figure}[h!] 
\centering
\includegraphics[width=0.65\textwidth,trim=0.35cm 1.5cm 1.5cm 1.2cm,clip]{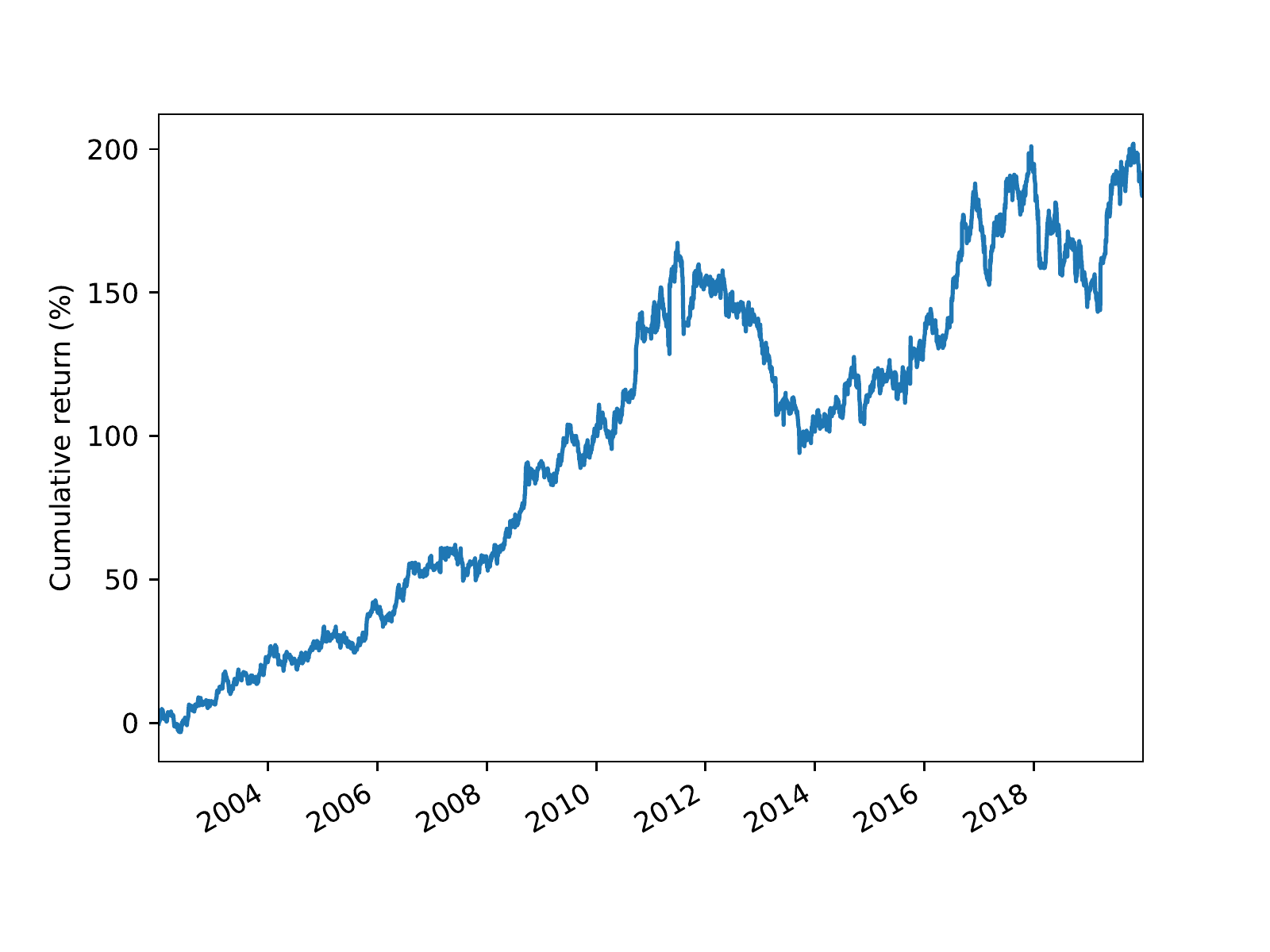}
\caption{Cumulative return for the financial forecasting signal; the signal is scaled to target a 10\% yearly volatility.}
	\label{fig:pnl_curve}
\end{figure}

The cumulative profit of the signal is displayed in   \Cref{fig:pnl_curve}.
The trading signal results in an annualised Sharpe Ratio of 0.62  with a corresponding significant one-sided p-value of $p < 0.004$ \citep{Opdyke2007}. We compare this with the Sharpe Ratio of 0.40 for the S\&P500 market return on the same period. Further, the trading signal exhibits a low correlation (0.04) with the market return. This suggests that the trading signal cannot be explained by market equilibrium returns. The mean daily return of the trading signal is $2.4$ basis points\footnote{Cf a mean daily market return of $3.0$ basis points.}.

We observe  in   \Cref{fig:pnl_curve} that there is a decay in the performance of the signal after 2012; this can be compared with the reduction in clustering persistence observed after 2012 in   \Cref{fig:yearly_ari}, and with the observation in the work of \cite{Curme2015} that the informational efficiency of the market appears to increase in 2012 relative to earlier years.

\paragraph{Ablation study}

We conduct an ablation study in order to test the importance of the lead-lag clustering structure on the observed performance of the trading signal. Specifically, under the null hypothesis that there is no lead-lag cluster structure in US equity returns, the clustering for the US equities is drawn uniformly at random from the set of permutations on cluster labels. Therefore, under the hypothesis of lead-lag cluster structure, the Sharpe Ratio of the trading signal described in \Cref{Signal construction} should be consistent with the distribution over the Sharpe Ratios of trading signals that are computed with permuted cluster labels. We use 200 Monte Carlo samples from the null distribution that computes the Sharpe Ratio of the same trading signal pipeline described in \Cref{Signal construction} but with any clustering in this pipeline drawn uniformly at random from $S_p$. Under the null hypothesis, the Monte Carlo probability that the Sharpe Ratio is greater than or equal to the observed Sharpe Ratio of $0.62$ is $1/201$. We thus reject the null hypothesis with p-value $p < 0.005$, and conclude that the lead-lag cluster structure is significant in the construction of the predictive trading signal.

A caveat to our results is that we do not take into account transaction costs when calculating the profit of our signal. These may be significant in practice given the basis point size of the average daily returns.
On the other hand, the turnover of the trading signal, which is based on a weekly smoothing of lagged returns, is relatively low.
Regardless of the economic significance of the signal, it is clear that the clustered lead-lag structure is statistically strong enough to be used as a predictive signal for equity returns.

\section{Conclusion} \label{sec:Conclusion}
We propose a methodology for the problem of data-driven detection of leading and lagging clusters of time series. Our unsupervised learning method can capture general, non-linear lead-lag correlations and leverages a state-of-the-art directed network clustering algorithm which is able to detect clusters with high flow imbalance. When applied to US equity data, our method produces a clustering that is statistically significant but that cannot be explained by three prominent lead-lag hypotheses in the empirical finance literature; this suggests that our methodology is a useful tool for the exploration of novel lead-lag mechanisms in the discipline of empirical finance.  Furthermore, we find that our method can be employed for challenging downstream forecasting tasks in noisy, high-dimensional settings. In particular, we show how our method can be used for the construction of a statistically significant, parsimonious trading signal in the US equity market.

In addition to the financial domain, the applicability of our proposed methodology extends to other areas -- such as economics, medicine and earth sciences -- that are characterised by large multivariate time series data which exhibit a latent lead-lag structure. Finally, our network approach to time series, which is able to infer global clustering structure based on local pairwise interactions, can be applied to general pairwise directed interaction data between time series variables. Thus, our framework may be generalised beyond \textit{lead-lag} interactions, in order to discover cluster structure in high-dimensional time-series systems based on \textit{general} directed interactions.

\bibliographystyle{abbrvnat}
\bibliography{bibliography.bib} 

\begin{thebibliography}{79}
\providecommand{\natexlab}[1]{#1}
\providecommand{\url}[1]{\texttt{#1}}
\expandafter\ifx\csname urlstyle\endcsname\relax
  \providecommand{\doi}[1]{doi: #1}\else
  \providecommand{\doi}{doi: \begingroup \urlstyle{rm}\Url}\fi

\bibitem[Asness et~al.(2013)Asness, Moskowitz, and Pedersen]{Asness2013}
C.~S. Asness, T.~J. Moskowitz, and L.~H. Pedersen.
\newblock {Value and Momentum Everywhere}.
\newblock \emph{Journal of Finance}, 68\penalty0 (3):\penalty0 929--985, 2013.
\newblock ISSN 00221082.
\newblock \doi{10.1111/jofi.12021}.

\bibitem[Badrinath et~al.(1995)Badrinath, Jayant, and Thomas]{Badrinath1995}
S.~G. Badrinath, R.~K. Jayant, and H.~N. Thomas.
\newblock {Of Shepards, Sheep and the cross-autocorrelations in equity
  returns}.
\newblock \emph{The Review of Financial Studies}, 8\penalty0 (2), 1995.

\bibitem[Basnarkov et~al.(2019)Basnarkov, Stojkoski, Utkovski, and
  Kocarev]{Basnarkov2019}
L.~Basnarkov, V.~Stojkoski, Z.~Utkovski, and L.~Kocarev.
\newblock {Lead-lag Relationships in Foreign Exchange Markets}.
\newblock \emph{arXiv}, \penalty0 (1906.10388v2), 2019.
\newblock \doi{10.1016/j.physa.2019.122986}.

\bibitem[Batson et~al.(2013)Batson, Spielman, Srivastava, and
  Teng]{graph_sparsification}
J.~Batson, D.~A. Spielman, N.~Srivastava, and S.-H. Teng.
\newblock Spectral sparsification of graphs: Theory and algorithms.
\newblock \emph{Commun. ACM}, 56\penalty0 (8):\penalty0 87–94, aug 2013.
\newblock ISSN 0001-0782.
\newblock \doi{10.1145/2492007.2492029}.

\bibitem[Biely and Thurner(2008)]{Biely2008}
C.~Biely and S.~Thurner.
\newblock {Random matrix ensembles of time-lagged correlation matrices:
  Derivation of eigenvalue spectra and analysis of financial time-series}.
\newblock \emph{Quantitative Finance}, 8\penalty0 (7):\penalty0 705--722, 2008.
\newblock ISSN 14697696.
\newblock \doi{10.1080/14697680701691477}.

\bibitem[Billio et~al.(2012)Billio, Getmansky, Lo, and Pelizzon]{Billio2012}
M.~Billio, M.~Getmansky, A.~W. Lo, and L.~Pelizzon.
\newblock {Econometric measures of connectedness and systemic risk in the
  finance and insurance sectors}.
\newblock \emph{Journal of Financial Economics}, 104\penalty0 (3):\penalty0
  535--559, 2012.
\newblock ISSN 0304405X.
\newblock \doi{10.1016/j.jfineco.2011.12.010}.

\bibitem[Bradley and Terry(1952)]{BradleyTerry1952}
R.~A. Bradley and M.~E. Terry.
\newblock Rank analysis of incomplete block designs: I. the method of paired
  comparisons.
\newblock \emph{Biometrika}, pages 324--345, 1952.

\bibitem[Brennan et~al.(1993)Brennan, Narasimhan, and Swaminathan]{Brennan1993}
M.~J.~. Brennan, J.~Narasimhan, and B.~Swaminathan.
\newblock {Investment Analysis and the Adjustment of Stock Prices to Common
  Information Source}.
\newblock \emph{The Review of Financial Studies}, 6\penalty0 (4):\penalty0
  799--824, 1993.

\bibitem[Camilleri et~al.(2019)Camilleri, Scicluna, and Bai]{CAMILLERI2019170}
S.~J. Camilleri, N.~Scicluna, and Y.~Bai.
\newblock Do stock markets lead or lag macroeconomic variables? evidence from
  select european countries.
\newblock \emph{The North American Journal of Economics and Finance},
  48:\penalty0 170--186, 2019.
\newblock ISSN 1062-9408.
\newblock \doi{https://doi.org/10.1016/j.najef.2019.01.019}.

\bibitem[Campbell et~al.(1997)Campbell, Lo, and MacKinlay]{Campbell1997}
J.~Y. Campbell, A.~W. Lo, and A.~C. MacKinlay.
\newblock {The econometrics of financial markets}.
\newblock pages 11, 74--78, 84, 128--132, 1997.
\newblock ISSN 0893-9454.
\newblock \doi{10.1515/9781400830213-004}.

\bibitem[Chau et~al.(2020)Chau, Cucuringu, and Sejdinovic]{chau2020spectral}
S.~L. Chau, M.~Cucuringu, and D.~Sejdinovic.
\newblock Spectral ranking with covariates.
\newblock \emph{arXiv preprint arXiv:2005.04035}, 2020.

\bibitem[Chevyrev and Kormilitzin(2016)]{Chevyreva}
I.~Chevyrev and A.~Kormilitzin.
\newblock {A Primer on the Signature Method in Machine Learning}.
\newblock \emph{arXiv}, \penalty0 (1603.03788v1), 2016.

\bibitem[Chordia and Swaminathan(2000)]{Chordia2000}
T.~Chordia and B.~Swaminathan.
\newblock {Trading Volume and Cross-Autocorrelations in Stock Returns}.
\newblock \emph{The Journal of Finance}, LV\penalty0 (2):\penalty0 913--935,
  2000.

\bibitem[Cohen and Frazzini(2008)]{Cohen2008}
L.~Cohen and A.~Frazzini.
\newblock {Economic links and predictable returns}.
\newblock \emph{Journal of Finance}, 63\penalty0 (4):\penalty0 1977--2011,
  2008.
\newblock ISSN 00221082.
\newblock \doi{10.1111/j.1540-6261.2008.01379.x}.

\bibitem[Conrad et~al.(1991)Conrad, Gultekin, and Kaul]{Conrad1991}
J.~Conrad, M.~Gultekin, and G.~Kaul.
\newblock {Asymmetric Predictability of Conditional Variances}.
\newblock \emph{The Review of Financial Studies}, 4\penalty0 (4):\penalty0
  597--622, 1991.

\bibitem[Cont(2001)]{Cont2001}
R.~Cont.
\newblock {Empirical properties of asset returns: Stylized facts and
  statistical issues}.
\newblock \emph{Quantitative Finance}, 1\penalty0 (2):\penalty0 223--236, 2001.
\newblock ISSN 14697696.
\newblock \doi{10.1080/713665670}.

\bibitem[Cucuringu(2016)]{syncRank}
M.~Cucuringu.
\newblock {Sync-Rank: Robust Ranking, Constrained Ranking and Rank Aggregation
  via Eigenvector and Semidefinite Programming Synchronization}.
\newblock \emph{IEEE Transactions on Network Science and Engineering},
  3\penalty0 (1):\penalty0 58--79, 2016.

\bibitem[Cucuringu et~al.(2020)Cucuringu, Li, Sun, and Zanetti]{Cucuringu2019}
M.~Cucuringu, H.~Li, H.~Sun, and L.~Zanetti.
\newblock {Hermitian matrices for clustering directed graphs: insights and
  applications}.
\newblock \emph{AISTATS}, \penalty0 (c):\penalty0 1--19, 2020.

\bibitem[Curme et~al.(2015{\natexlab{a}})Curme, Tumminello, Mantegna, Stanley,
  and Kenett]{Curme2015}
C.~Curme, M.~Tumminello, R.~N. Mantegna, H.~E. Stanley, and D.~Y. Kenett.
\newblock {Emergence of statistically validated financial intraday lead-lag
  relationships}.
\newblock \emph{Quantitative Finance}, 15\penalty0 (8):\penalty0 1375--1386,
  2015{\natexlab{a}}.
\newblock ISSN 14697696.
\newblock \doi{10.1080/14697688.2015.1032545}.

\bibitem[Curme et~al.(2015{\natexlab{b}})Curme, Tumminello, Mantegna, Stanley,
  and Kenett]{Curme2015a}
C.~Curme, M.~Tumminello, R.~N. Mantegna, H.~E. Stanley, and D.~Y. Kenett.
\newblock {How Lead-Lag Correlations Affect the Intraday Pattern of Collective
  Stock Dynamics}.
\newblock 2015{\natexlab{b}}.

\bibitem[d'Aspremont et~al.(2021)d'Aspremont, Cucuringu, and
  Tyagi]{SVDRankSync_JMLR}
A.~d'Aspremont, M.~Cucuringu, and H.~Tyagi.
\newblock Ranking and synchronization from pairwise measurements via svd.
\newblock \emph{Journal of Machine Learning Research}, 22\penalty0
  (19):\penalty0 1--63, 2021.
\newblock URL \url{http://jmlr.org/papers/v22/19-542.html}.

\bibitem[De~Bacco et~al.(2018)De~Bacco, Larremore, and
  Moore]{CaterinaDeBacco_Ranking}
C.~De~Bacco, D.~B. Larremore, and C.~Moore.
\newblock A physical model for efficient ranking in networks.
\newblock \emph{Science Advances}, 4\penalty0 (7), 2018.

\bibitem[Dugu{\'{e}} and Perez(2015)]{Dugue2015}
N.~Dugu{\'{e}} and A.~Perez.
\newblock {Directed Louvain : maximizing modularity in directed networks}.
\newblock \emph{HAL archives ouvertes}, pages 0--14, 2015.
\newblock URL \url{https://hal.archives-ouvertes.fr/hal-01231784}.

\bibitem[Fama and French(1993)]{Fama1993}
E.~F. Fama and K.~R. French.
\newblock {Common risk factors in the returns on stocks and bonds}.
\newblock \emph{Journal of Financial Economics}, 33\penalty0 (1):\penalty0
  3--56, 1993.
\newblock ISSN 0046-9777.
\newblock \doi{10.2469/dig.v36.n3.4225}.

\bibitem[Farrell(1974)]{Farrell1974}
J.~Farrell.
\newblock {Analyzing Covariation of Returns to Determine Homogeneous Stock
  Groupings}.
\newblock \emph{Journal of Business}, 47\penalty0 (2):\penalty0 186--207, 1974.

\bibitem[Fiedor(2014)]{Fiedor2014}
P.~Fiedor.
\newblock {Information-theoretic approach to lead-lag effect on financial
  markets}.
\newblock \emph{European Physical Journal B}, 87\penalty0 (8), 2014.
\newblock ISSN 14346036.
\newblock \doi{10.1140/epjb/e2014-50108-3}.

\bibitem[Fogel et~al.(2016)Fogel, d'Aspremont, and Vojnovic]{fogel2016spectral}
F.~Fogel, A.~d'Aspremont, and M.~Vojnovic.
\newblock Spectral ranking using seriation.
\newblock \emph{Journal of Machine Learning Research}, 17\penalty0
  (88):\penalty0 1--45, 2016.

\bibitem[Gleich and Lim(2011)]{gleich2011rank}
D.~F. Gleich and L.-h. Lim.
\newblock Rank aggregation via nuclear norm minimization.
\newblock In \emph{Proceedings of the 17th ACM SIGKDD international conference
  on Knowledge discovery and data mining}, pages 60--68. ACM, 2011.

\bibitem[Google(2012)]{Kumar2012}
Google.
\newblock {The PageRank Citation Ranking: Bringing Order to the Web January}.
\newblock \emph{Proceedings - 2012 IEEE International Symposium on Workload
  Characterization, IISWC 2012}, pages 111--112, 2012.
\newblock \doi{10.1109/IISWC.2012.6402911}.

\bibitem[Gretton et~al.(2012)Gretton, Borgwardt, Rasch, Sch{\"{o}}lkopf, and
  Smola]{Gretton2012}
A.~Gretton, K.~M. Borgwardt, M.~J. Rasch, B.~Sch{\"{o}}lkopf, and A.~Smola.
\newblock {A kernel two-sample test}.
\newblock \emph{Journal of Machine Learning Research}, 13:\penalty0 723--773,
  2012.
\newblock ISSN 15324435.

\bibitem[Gyurk{\'{o}} et~al.(2014)Gyurk{\'{o}}, Lyons, Kontkowski, and
  Field]{Gyurko2014}
L.~G. Gyurk{\'{o}}, T.~Lyons, M.~Kontkowski, and J.~Field.
\newblock {Extracting information from the signature of a financial data
  stream}.
\newblock \emph{arXiv}, pages 1--22, 2014.

\bibitem[Harzallah and Sadourny(1997)]{Harzallah1997}
A.~Harzallah and R.~Sadourny.
\newblock {Observed lead-lag relationships between Indian summer monsoon and
  some meteorological variables}.
\newblock \emph{Climate Dynamics}, 13\penalty0 (9):\penalty0 635--648, 1997.
\newblock ISSN 14320894.
\newblock \doi{10.1007/s003820050187}.

\bibitem[He et~al.(2021)He, Reinert, and Cucuringu]{he2021digrac}
Y.~He, G.~Reinert, and M.~Cucuringu.
\newblock Digrac: Digraph clustering with flow imbalance.
\newblock 2021.

\bibitem[Hu and Lau(2013)]{Graph_Sampling_Hu}
P.~Hu and W.~C. Lau.
\newblock A survey and taxonomy of graph sampling, 2013.

\bibitem[Huber(1962)]{Huber1962}
P.~J. Huber.
\newblock {Pairwise Comparison and Ranking: Optimum Properties of the Row Sum
  Procedure}.
\newblock \emph{The Annals of Mathematical Statistics}, 1962.

\bibitem[Huth(2012)]{Huth2012}
N.~Huth.
\newblock {High Frequency Lead / lag Relationships Empirical facts}.
\newblock \emph{Journal of Empirical Finance}, 26\penalty0 (March
  2014):\penalty0 41--58, 2012.
\newblock URL
  \url{http://www.sciencedirect.com/science/article/pii/S0927539814000048}.

\bibitem[Iyetomi et~al.(2020)Iyetomi, Aoyama, Fujiwara, Souma, Vodenska, and
  Yoshikawa]{iyetomi2020relationship}
H.~Iyetomi, H.~Aoyama, Y.~Fujiwara, W.~Souma, I.~Vodenska, and H.~Yoshikawa.
\newblock Relationship between macroeconomic indicators and economic cycles in
  us.
\newblock \emph{Scientific reports}, 10\penalty0 (1):\penalty0 1--12, 2020.

\bibitem[Janzing et~al.(2013)Janzing, Balduzzi, Grosse-Wentrup, and
  Sch{\"{o}}lkopf]{Janzing2013}
D.~Janzing, D.~Balduzzi, M.~Grosse-Wentrup, and B.~Sch{\"{o}}lkopf.
\newblock {Quantifying causal influences}.
\newblock \emph{Annals of Statistics}, 41\penalty0 (5):\penalty0 2324--2358,
  2013.
\newblock ISSN 00905364.
\newblock \doi{10.1214/13-AOS1145}.

\bibitem[Jegadeesh and Titman(1995)]{Jegadeesh1995}
N.~Jegadeesh and S.~Titman.
\newblock {Overreaction , Delayed Reaction , and Contrarian Profits}.
\newblock \emph{The Review of Financial Studies}, 8\penalty0 (4):\penalty0
  973--993, 1995.

\bibitem[Kendall(1938)]{Kendall1938}
M.~G. Kendall.
\newblock {A New Measure of Rank Correlation}.
\newblock \emph{Biometrika}, 30\penalty0 (1):\penalty0 81--93, 1938.

\bibitem[Laenen and Sun(2020)]{Laenen2020}
S.~Laenen and H.~Sun.
\newblock {Higher-order spectral clustering of directed graphs}.
\newblock \emph{Advances in Neural Information Processing Systems},
  2020\penalty0 (NeurIPS), 2020.
\newblock ISSN 10495258.

\bibitem[Levin et~al.(2016)Levin, Lyons, and Ni]{Levin2013a}
D.~Levin, T.~Lyons, and H.~Ni.
\newblock {Learning from the past, predicting the statistics for the future,
  learning an evolving system}.
\newblock \emph{arXiv}, \penalty0 (291244):\penalty0 1--40, 2016.

\bibitem[Liao et~al.(2014)Liao, Huang, Shi, and Jin]{Liao2014}
C.~Liao, Y.~Huang, X.~Shi, and X.~Jin.
\newblock {Mining influence in evolving entities: A study on stock market}.
\newblock \emph{DSAA 2014 - Proceedings of the 2014 IEEE International
  Conference on Data Science and Advanced Analytics}, pages 244--250, 2014.
\newblock \doi{10.1109/DSAA.2014.7058080}.

\bibitem[Lin et~al.(2013)Lin, Ding, Yan, Yu, and Giua]{Lin2013LeaderfollowerFV}
Z.~Lin, W.~Ding, G.~Yan, C.~Yu, and A.~Giua.
\newblock Leader-follower formation via complex laplacian.
\newblock \emph{Autom.}, 49:\penalty0 1900--1906, 2013.

\bibitem[Lo and MacKinlay(1990)]{Lo1990}
A.~W. Lo and A.~C. MacKinlay.
\newblock {When are Contrarian Profits Due to Stock Market Overreaction}.
\newblock \emph{The Review of Financial Studies}, 3\penalty0 (2):\penalty0
  175--205, 1990.

\bibitem[Malkiel and Fama(1970)]{Malkiel1970}
B.~G. Malkiel and E.~Fama.
\newblock {Efficient Capital Markets: A Review of Theory and Empirical Work}.
\newblock \emph{The Journal of Finance}, 25\penalty0 (2), 1970.
\newblock ISSN 00221082.
\newblock \doi{10.2307/2325488}.

\bibitem[Marti et~al.(2016)Marti, Andler, Nielsen, and Donnat]{Marti2016}
G.~Marti, S.~Andler, F.~Nielsen, and P.~Donnat.
\newblock {Exploring and measuring non-linear correlations: Copulas, Lightspeed
  Transportation and Clustering}.
\newblock \emph{arXiv}, \penalty0 (1610.09659v1), 2016.

\bibitem[Marti et~al.(2019)Marti, Nielsen, Bi{\'{n}}kowski, and
  Donnat]{Marti2017}
G.~Marti, F.~Nielsen, M.~Bi{\'{n}}kowski, and P.~Donnat.
\newblock {A review of two decades of correlations, hierarchies, networks and
  clustering in financial markets}.
\newblock \emph{arXiv}, \penalty0 (1703.00485v5):\penalty0 1--34, 2019.

\bibitem[Menzly and Ozbas(2010)]{Menzly2010}
L.~Menzly and O.~Ozbas.
\newblock {Market segmentation and cross-predictability of returns}.
\newblock \emph{Journal of Finance}, 65\penalty0 (4):\penalty0 1555--1580,
  2010.
\newblock ISSN 00221082.
\newblock \doi{10.1111/j.1540-6261.2010.01578.x}.

\bibitem[Namaki et~al.(2011)Namaki, Shirazi, Raei, and Jafari]{Namaki2011a}
A.~Namaki, A.~H. Shirazi, R.~Raei, and G.~R. Jafari.
\newblock {Network analysis of a financial market based on genuine correlation
  and threshold method}.
\newblock \emph{Physica A: Statistical Mechanics and its Applications},
  390\penalty0 (21-22):\penalty0 3835--3841, 2011.
\newblock ISSN 03784371.
\newblock \doi{10.1016/j.physa.2011.06.033}.

\bibitem[Newman(2018)]{newman2018networks}
M.~Newman.
\newblock \emph{Networks}.
\newblock Oxford University Press, 2nd edition, 2018.

\bibitem[Opdyke(2007)]{Opdyke2007}
J.~D. Opdyke.
\newblock {Comparing Sharpe ratios: So where are the p-values?}
\newblock \emph{Journal of Asset Management}, 8\penalty0 (5):\penalty0
  308--336, 2007.
\newblock ISSN 1470-8272.
\newblock \doi{10.1057/palgrave.jam.2250084}.

\bibitem[Page et~al.(1998)Page, Brin, Motwani, and Winograd]{Pageetal98}
L.~Page, S.~Brin, R.~Motwani, and T.~Winograd.
\newblock {The PageRank citation ranking: Bringing order to the Web}.
\newblock In \emph{Proceedings of the 7th International World Wide Web
  Conference}, pages 161--172, 1998.

\bibitem[Pentney and Meila(2005)]{Pentney2005}
W.~Pentney and M.~Meila.
\newblock {Spectral clustering of biological sequence data}.
\newblock \emph{Proceedings of the National Conference on Artificial
  Intelligence}, 2:\penalty0 845--850, 2005.

\bibitem[Podobnik et~al.(2010)Podobnik, Wang, Horvatic, Grosse, and
  Stanley]{Podobnik2010a}
B.~Podobnik, D.~Wang, D.~Horvatic, I.~Grosse, and H.~E. Stanley.
\newblock {Time-lag cross-correlations in collective phenomena}.
\newblock \emph{EPL}, 90\penalty0 (68001), 2010.
\newblock ISSN 02955075.
\newblock \doi{10.1209/0295-5075/90/68001}.

\bibitem[Reizenstein and Graham(2018)]{Reizenstein2018}
J.~Reizenstein and B.~Graham.
\newblock {The iisignature library: efficient calculation of iterated-integral
  signatures and log signatures}.
\newblock \emph{arXiv}, 1802.08252:\penalty0 1--18, 2018.

\bibitem[Reshef et~al.(2011)Reshef, Reshef, Finucane, Grossman, McVean,
  Turnbaugh, Lander, Mitzenmacher, and Sabeti]{Reshef2011}
D.~N. Reshef, Y.~A. Reshef, H.~K. Finucane, S.~R. Grossman, G.~McVean, P.~J.
  Turnbaugh, E.~S. Lander, M.~Mitzenmacher, and P.~C. Sabeti.
\newblock {Detecting Novel Associations in Large Datasets}.
\newblock \emph{Science}, 334\penalty0 (6062):\penalty0 1518--1524, 2011.
\newblock \doi{10.1126/science.1205438.Detecting}.

\bibitem[Rohe et~al.(2016)Rohe, Qin, and Yu]{Rohe2016}
K.~Rohe, T.~Qin, and B.~Yu.
\newblock {Co-clustering directed graphs to discover asymmetries and
  directional communities}.
\newblock \emph{Proceedings of the National Academy of Sciences of the United
  States of America}, 113\penalty0 (45):\penalty0 12679--12684, 2016.
\newblock ISSN 10916490.
\newblock \doi{10.1073/pnas.1525793113}.

\bibitem[Runge et~al.(2019)Runge, Nowack, Kretschmer, Flaxman, and
  Sejdinovic]{Runge2017}
J.~Runge, P.~Nowack, M.~Kretschmer, S.~Flaxman, and D.~Sejdinovic.
\newblock {Detecting causal associations in large nonlinear time series
  datasets}.
\newblock \emph{Science Advances}, 5\penalty0 (11), 2019.
\newblock ISSN 23318422.

\bibitem[Sandoval(2014)]{Sandoval2014a}
L.~Sandoval.
\newblock {Structure of a Global Network of financial companies based on
  transfer entropy}.
\newblock \emph{Entropy}, 16\penalty0 (8):\penalty0 4443--4482, 2014.
\newblock ISSN 10994300.
\newblock \doi{10.3390/e16084443}.

\bibitem[Sandoval and Franca(2012)]{Sandoval2012}
L.~Sandoval and I.~D.~P. Franca.
\newblock {Correlation of financial markets in times of crisis}.
\newblock \emph{Physica A: Statistical Mechanics and its Applications},
  391\penalty0 (1-2):\penalty0 187--208, 2012.
\newblock ISSN 03784371.
\newblock \doi{10.1016/j.physa.2011.07.023}.

\bibitem[Satuluri and Parthasarathy(2011)]{Satuluri2011}
V.~Satuluri and S.~Parthasarathy.
\newblock {Symmetrizations for clustering directed graphs}.
\newblock \emph{ACM International Conference Proceeding Series}, \penalty0
  (i):\penalty0 343--354, 2011.
\newblock \doi{10.1145/1951365.1951407}.

\bibitem[Scherbina and Schlusche(2015)]{Scherbina2013}
A.~D. Scherbina and B.~Schlusche.
\newblock Cross-firm information flows and the predictability of stock returns.
\newblock \emph{SSRN Electronic Journal}, 2015.

\bibitem[Service(2020)]{WRDS}
W.~R.~D. Service.
\newblock Center for research in security prices (crsp).
\newblock 2020.

\bibitem[Shi and Malik(2000)]{Shi2000}
J.~Shi and J.~Malik.
\newblock {Normalized Cuts and Image Segmentation}.
\newblock \emph{IEEE Transactions on Pattern Analysis and Machine
  Intelligence}, 22\penalty0 (8), 2000.
\newblock \doi{10.1109/ICIP.2014.7025680}.

\bibitem[Shojaie and Fox(2021)]{Shojaie2021}
A.~Shojaie and E.~B. Fox.
\newblock {Granger Causality: A Review and Recent Advances}.
\newblock \emph{arXiv}, 2105.02675, 2021.

\bibitem[Sornette and Zhou(2005)]{Sornette2005}
D.~Sornette and W.~X. Zhou.
\newblock {Non-parametric determination of real-time lag structure between two
  time series: The 'optimal thermal causal path' method}.
\newblock \emph{Quantitative Finance}, 5\penalty0 (6):\penalty0 577--591, 2005.
\newblock ISSN 14697688.
\newblock \doi{10.1080/14697680500383763}.

\bibitem[Stavroglou et~al.(2017)Stavroglou, Pantelous, Soramaki, and
  Zuev]{Stavroglou2017}
S.~Stavroglou, A.~Pantelous, K.~Soramaki, and K.~Zuev.
\newblock {Causality networks of financial assets}.
\newblock \emph{The Journal of Network Theory in Finance}, 3\penalty0
  (2):\penalty0 17--67, 2017.
\newblock ISSN 20557795.
\newblock \doi{10.21314/jntf.2017.029}.

\bibitem[Sussman et~al.(2012)Sussman, Tang, Fishkind, and Priebe]{Sussman2012}
D.~L. Sussman, M.~Tang, D.~E. Fishkind, and C.~E. Priebe.
\newblock {A consistent adjacency spectral embedding for stochastic blockmodel
  graphs}.
\newblock \emph{Journal of the American Statistical Association}, 107\penalty0
  (499):\penalty0 1119--1128, 2012.
\newblock ISSN 01621459.
\newblock \doi{10.1080/01621459.2012.699795}.

\bibitem[Sz{\'{e}}kely et~al.(2007)Sz{\'{e}}kely, Rizzo, and
  Bakirov]{Szekely2007}
G.~J. Sz{\'{e}}kely, M.~L. Rizzo, and N.~K. Bakirov.
\newblock {Measuring and testing dependence by correlation of distances}.
\newblock \emph{Annals of Statistics}, 35\penalty0 (6):\penalty0 2769--2794,
  2007.
\newblock ISSN 00905364.
\newblock \doi{10.1214/009053607000000505}.

\bibitem[Thorp(2011)]{Thorp2011}
E.~O. Thorp.
\newblock {The Kelly Criterion in Blackjack Sports Betting, and the Stock
  Market}.
\newblock In \emph{The Kelly Capital Growth Investment Criterion (Chapter 9)}.
  2011.
\newblock \doi{10.1016/s1872-0978(06)01009-x}.

\bibitem[Traag et~al.(2019)Traag, Waltman, and van Eck]{Traag2019}
V.~A. Traag, L.~Waltman, and N.~J. van Eck.
\newblock {From Louvain to Leiden: guaranteeing well-connected communities}.
\newblock \emph{Scientific Reports}, 9:\penalty0 1--12, 2019.
\newblock ISSN 20452322.
\newblock \doi{10.1038/s41598-019-41695-z}.

\bibitem[Tumminello et~al.(2010)Tumminello, Lillo, and
  Mantegna]{Tumminello2010}
M.~Tumminello, F.~Lillo, and R.~N. Mantegna.
\newblock {Correlation, hierarchies, and networks in financial markets}.
\newblock \emph{Journal of Economic Behavior and Organization}, 75\penalty0
  (1):\penalty0 40--58, 2010.
\newblock ISSN 01672681.
\newblock \doi{10.1016/j.jebo.2010.01.004}.

\bibitem[Underwood et~al.(2020)Underwood, Elliott, and
  Cucuringu]{MotifSpectralDirectedClustering_ANS_2020}
W.~G. Underwood, A.~Elliott, and M.~Cucuringu.
\newblock Motif-based spectral clustering of weighted directed networks.
\newblock \emph{Applied Network Science}, 5\penalty0 (62), September 2020.

\bibitem[V{\'{y}}rost et~al.(2015)V{\'{y}}rost, Ly{\'{o}}csa, and
  Baum{\"{o}}hl]{Vyrost2015}
T.~V{\'{y}}rost, {\v{S}}.~Ly{\'{o}}csa, and E.~Baum{\"{o}}hl.
\newblock {Granger causality stock market networks: Temporal proximity and
  preferential attachment}.
\newblock \emph{Physica A: Statistical Mechanics and its Applications},
  427:\penalty0 262--276, 2015.
\newblock ISSN 03784371.
\newblock \doi{10.1016/j.physa.2015.02.017}.

\bibitem[Wang et~al.(2017{\natexlab{a}})Wang, Tu, Chang, and Li]{Wang2017}
D.~Wang, J.~Tu, X.~Chang, and S.~Li.
\newblock {The lead–lag relationship between the spot and futures markets in
  China}.
\newblock \emph{Quantitative Finance}, 17\penalty0 (9):\penalty0 1447--1456,
  2017{\natexlab{a}}.
\newblock ISSN 14697696.
\newblock \doi{10.1080/14697688.2016.1264616}.

\bibitem[Wang et~al.(2017{\natexlab{b}})Wang, Xie, He, and Stanley]{Wang2017a}
G.~J. Wang, C.~Xie, K.~He, and H.~E. Stanley.
\newblock {Extreme risk spillover network: application to financial
  institutions}.
\newblock \emph{Quantitative Finance}, 17\penalty0 (9):\penalty0 1417--1433,
  2017{\natexlab{b}}.
\newblock ISSN 14697696.
\newblock \doi{10.1080/14697688.2016.1272762}.

\bibitem[Wu et~al.(2010)Wu, Ke, Yu, Yu, and Chen]{Wu2010}
D.~Wu, Y.~Ke, J.~X. Yu, P.~S. Yu, and L.~Chen.
\newblock {Detecting leaders from correlated time series}.
\newblock \emph{International Conference on Database Systems for Advanced
  Applications}, 5981 LNCS:\penalty0 352--367, 2010.
\newblock ISSN 03029743.
\newblock \doi{10.1007/978-3-642-12026-8_28}.

\bibitem[Xia et~al.(2018)Xia, You, Jiang, and Chen]{Xia2018}
L.~Xia, D.~You, X.~Jiang, and W.~Chen.
\newblock {Emergence and temporal structure of Lead–Lag correlations in
  collective stock dynamics}.
\newblock \emph{Physica A: Statistical Mechanics and its Applications},
  502:\penalty0 545--553, 2018.
\newblock ISSN 03784371.
\newblock \doi{10.1016/j.physa.2018.02.112}.

\end{thebibliography}

\appendix  

\section{Additional numerical experiments} \label{sec:Appendix}

\subsection{Synthetic data experiment: lead-lag results} \label{sec:Appendix - Synthetic data experiment: lead-lag results}

\begin{figure}[h!] 
	\centering
		\includegraphics[width=1\textwidth]{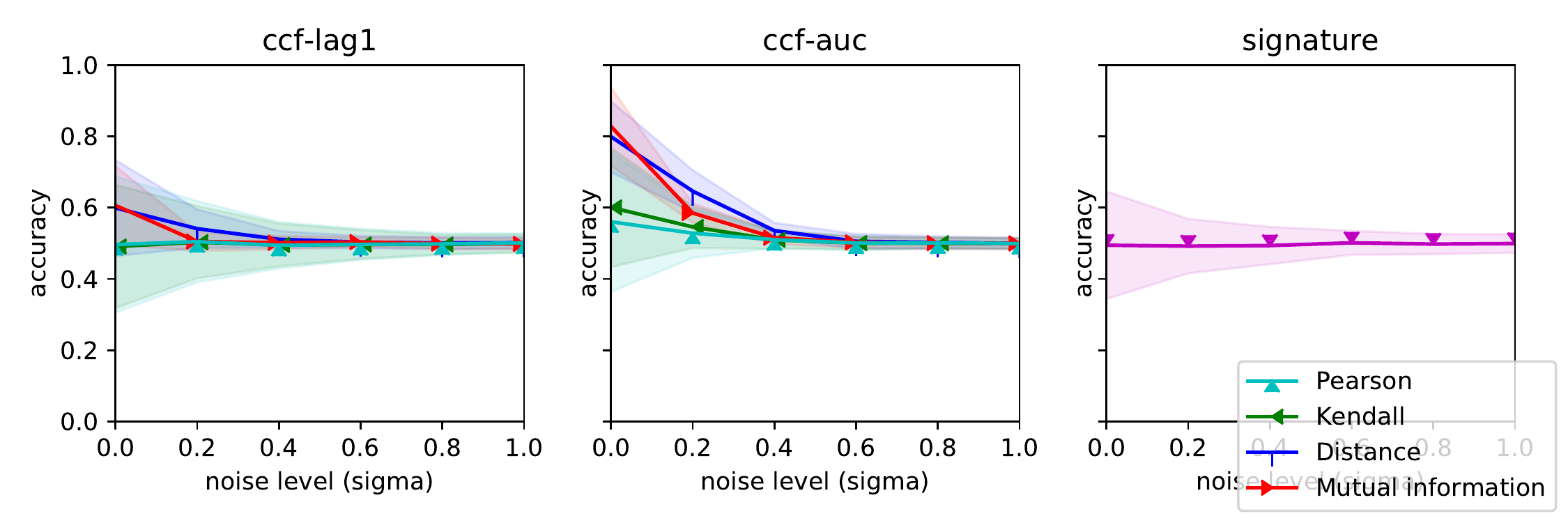}
	\caption{Average and confidence interval for accuracy by lead-lag detection method in the Legendre setting \eqref{eq:synLegendre}.}
	\label{fig:accuracy-legendre}
\end{figure}

\begin{figure}[h!] 
	\centering
		\includegraphics[width=1\textwidth]{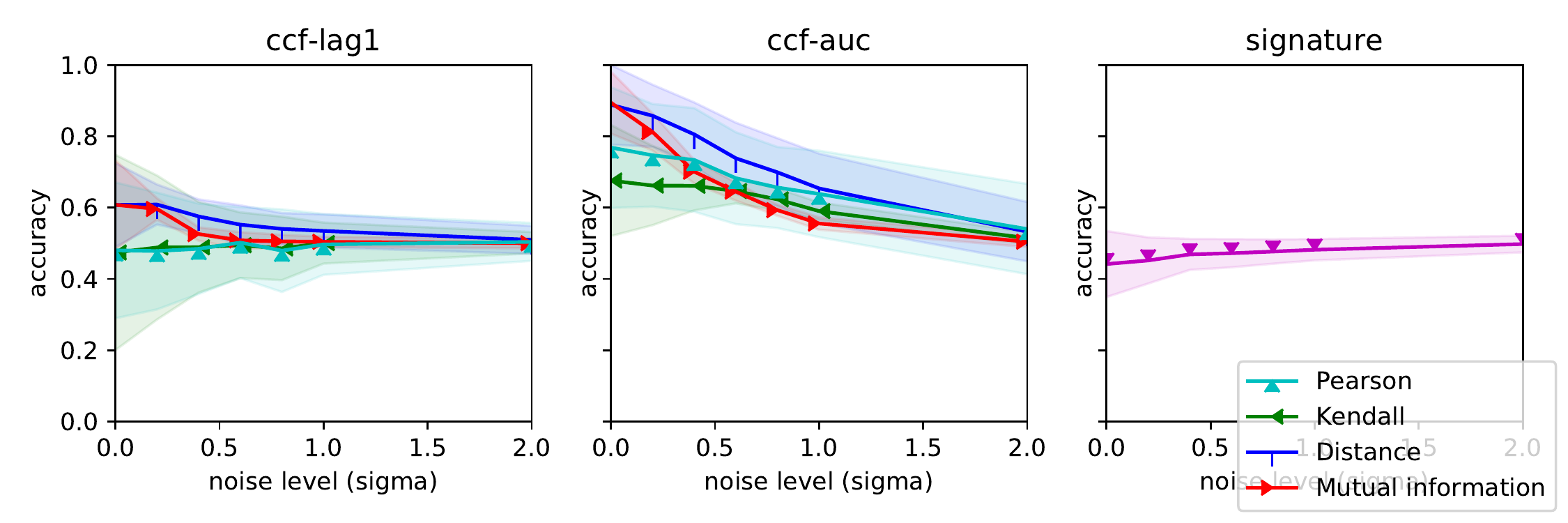}
	\caption{Average and confidence interval for accuracy by lead-lag detection method in the Hermite setting \eqref{eq:synHermite}. }
	\label{fig:accuracy-hermite}
\end{figure}

Figures \ref{fig:accuracy-legendre} and \ref{fig:accuracy-hermite}  display the lead-lag metric accuracy for the Legendre \eqref{eq:synLegendre} and Hermite \eqref{eq:synHermite}  synthetic data generating settings, respectively. We observe that the $\textbf{ccf-auc}$ method with the distance correlation performs best in these non-linear settings.

\subsection{Synthetic data experiment: clustering results} \label{sec:Appendix - Synthetic data experiment: clustering results}

\begin{figure}[h!] 
	\centering
		\includegraphics[width=0.55\textwidth]{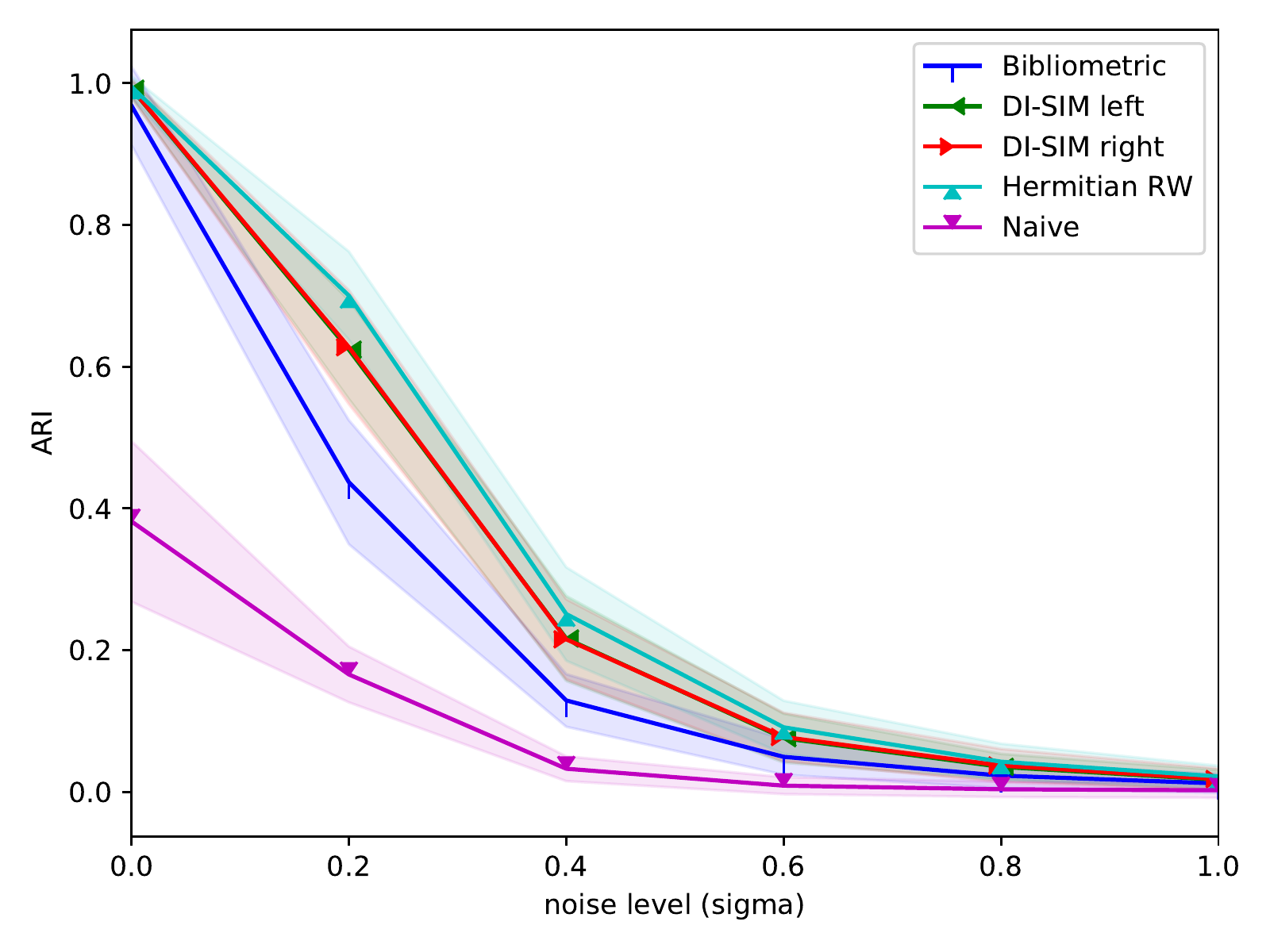}
	\caption{Average and confidence interval for the ARI by clustering method in the Legendre setting.}
	\label{fig:ari-legendre}
\end{figure}

\begin{figure}[h!] 
	\centering
		\includegraphics[width=0.55\textwidth]{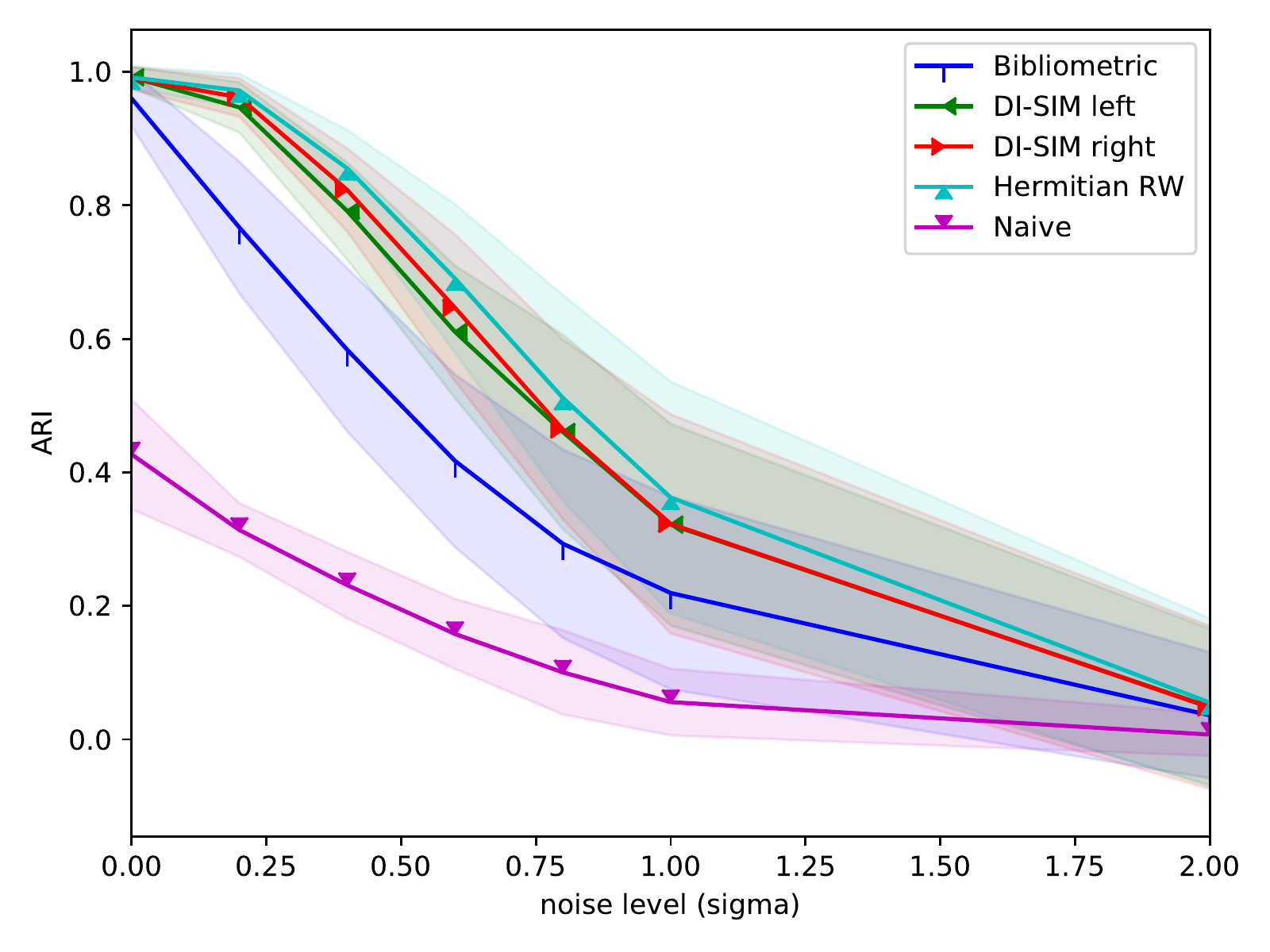}
	\caption{Average and confidence interval for the ARI by clustering method in the Hermite setting.}
	\label{fig:ari-hermite}
\end{figure}

\begin{figure}[h!] 
	\centering
		\includegraphics[width=0.55\textwidth]{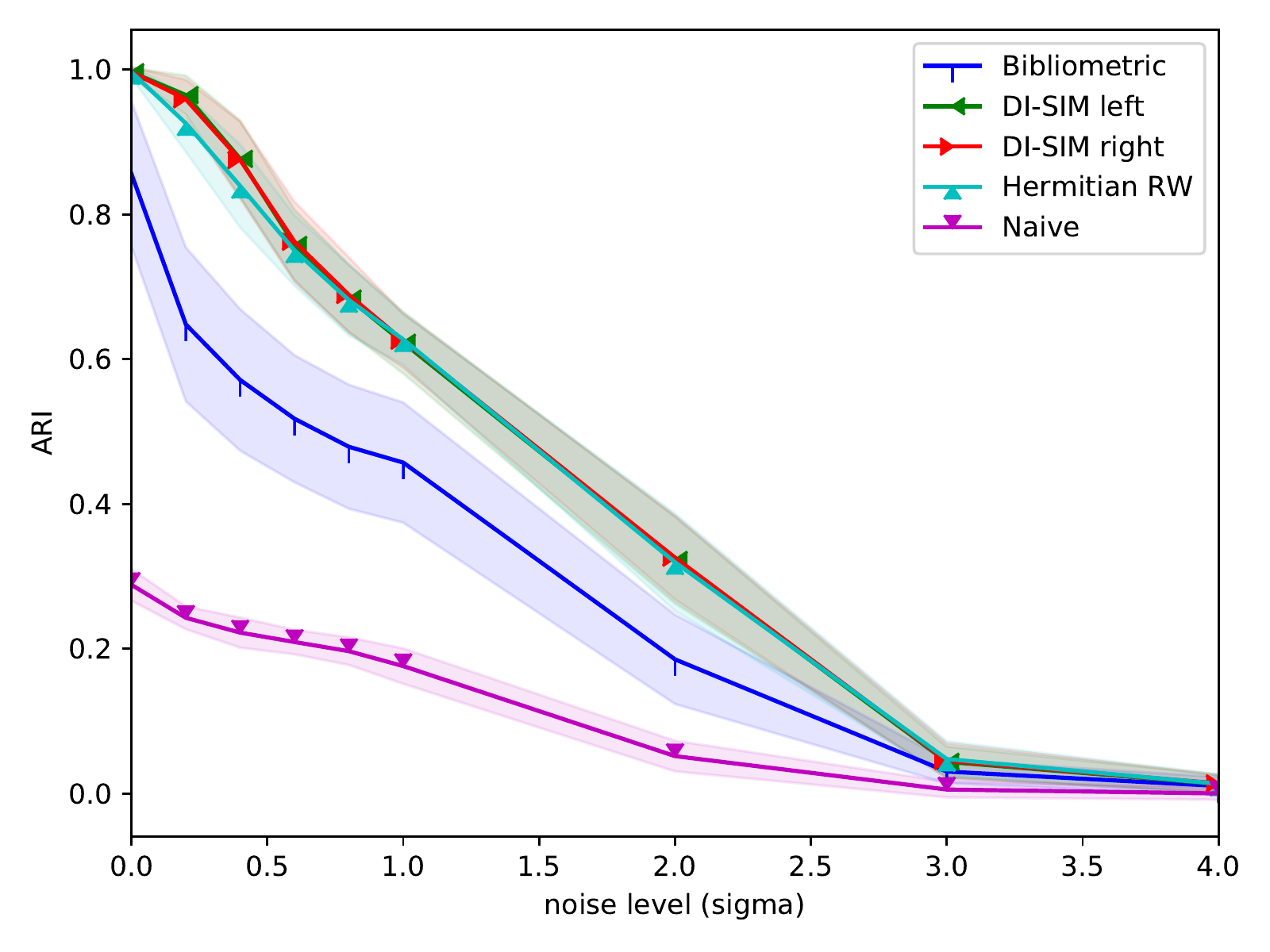}
	\caption{Average and confidence interval for the ARI by clustering method in the heterogeneous setting.}
	\label{fig:ari-heterogeneous}
\end{figure}

Figures \ref{fig:ari-legendre}, \ref{fig:ari-hermite} and \ref{fig:ari-heterogeneous}  display the ARI of our pipeline in the  Legendre  \eqref{eq:synLegendre}, Hermite \eqref{eq:synHermite} and Heterogeneous  \eqref{eq:synHetero} synthetic data generating settings, respectively. The pipeline performs best on average using the Hermitian RW clustering component in these settings. 

\subsection{Synthetic data experiment: interaction of lead-lag and clustering components} \label{sec: Appendix - Synthetic data experiment: interaction of lead-lag and clustering components}

\begin{figure}[h!] 
	\centering
		\includegraphics[width=1.\textwidth]{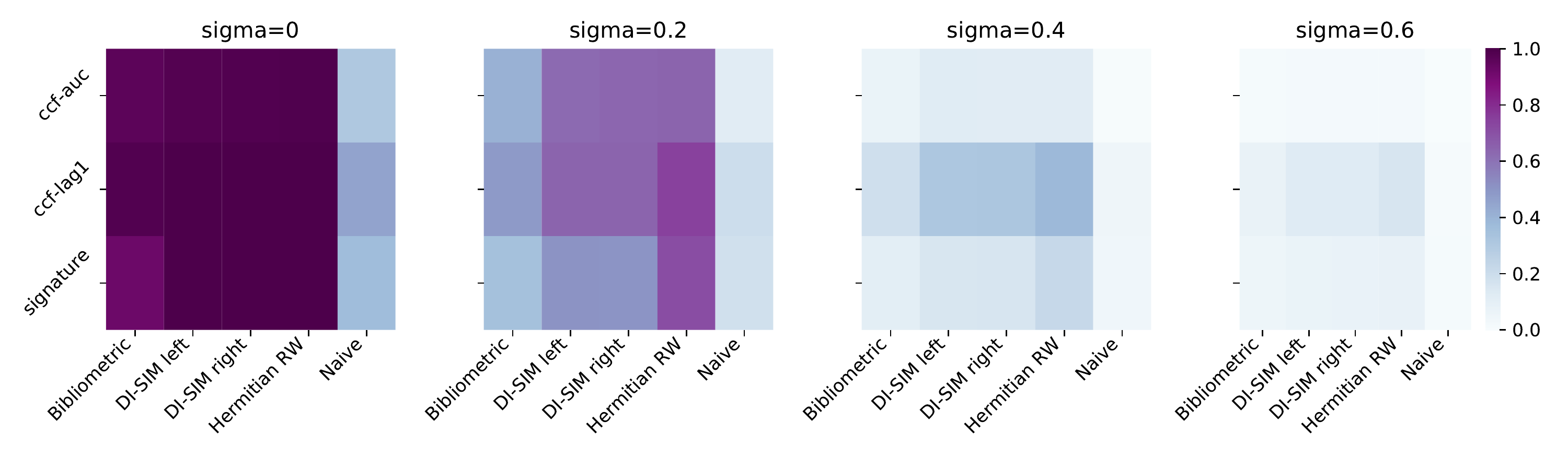}
	\caption{Average ARI by lead-lag and clustering component in the Legendre setting.}
	\label{fig:interaction_legendre}
\end{figure}

\begin{figure}[h!]
\centering
\includegraphics[width=1.\textwidth]{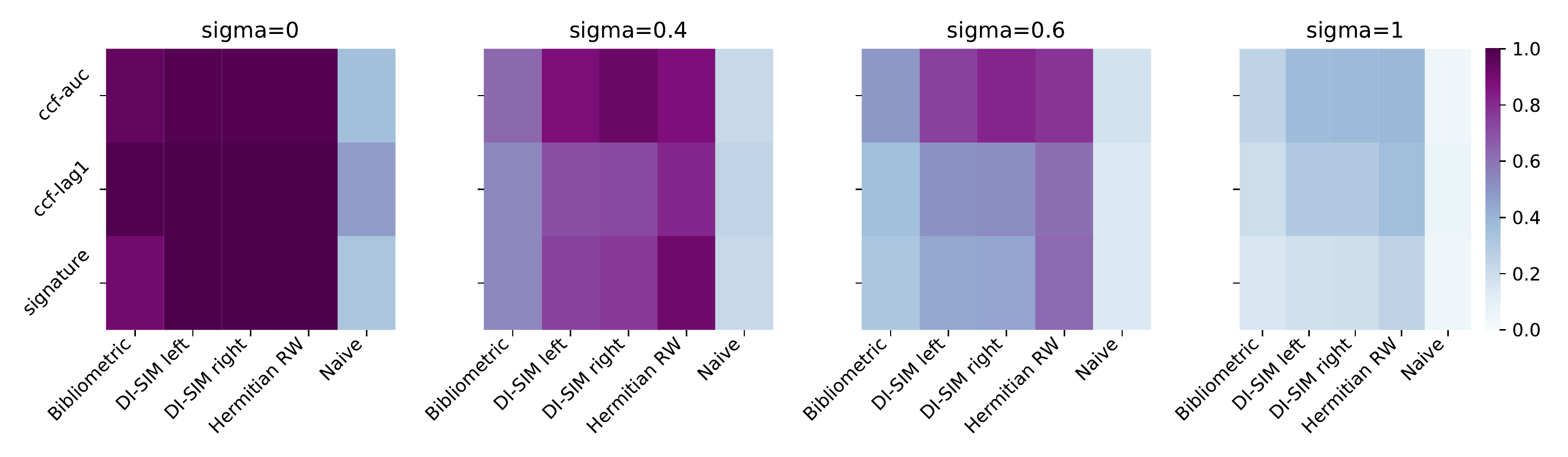}
\caption{Average ARI by lead-lag and clustering method in the Hermite setting.} 
\label{fig:interaction_hermite}
\end{figure}

\begin{figure}[h!]
	\centering
		\includegraphics[width=1.\textwidth]{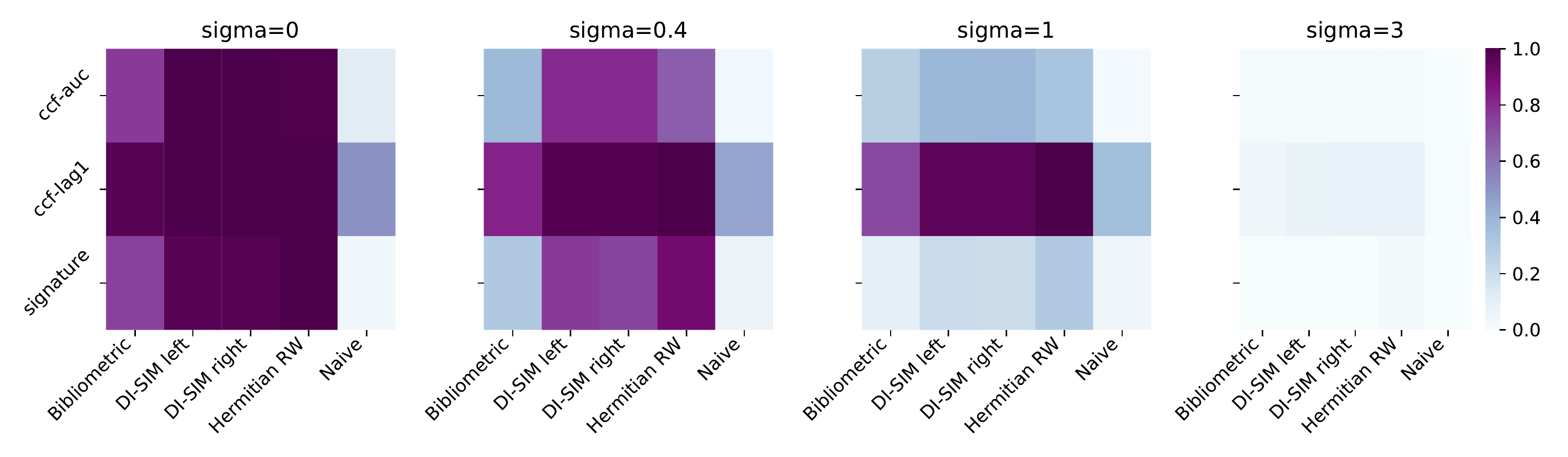}
	\caption{Average ARI by lead-lag and clustering method in the heterogeneous setting.} 
	\label{fig:interaction_heterogeneous}
\end{figure}

Figures \ref{fig:interaction_legendre}, \ref{fig:interaction_hermite} and \ref{fig:interaction_heterogeneous} display the ARI of the pipeline for each choice of lead-lag extraction and clustering components.

\subsection{Synthetic data ablation study: varying the hyperparameter corresponding to the number of clusters} \label{sec:Appendix - Synthetic data ablation study: varying hyperparameter corresponding to the number of clusters}

\begin{figure}[h!] 
	\centering
		\includegraphics[width=0.55\textwidth]{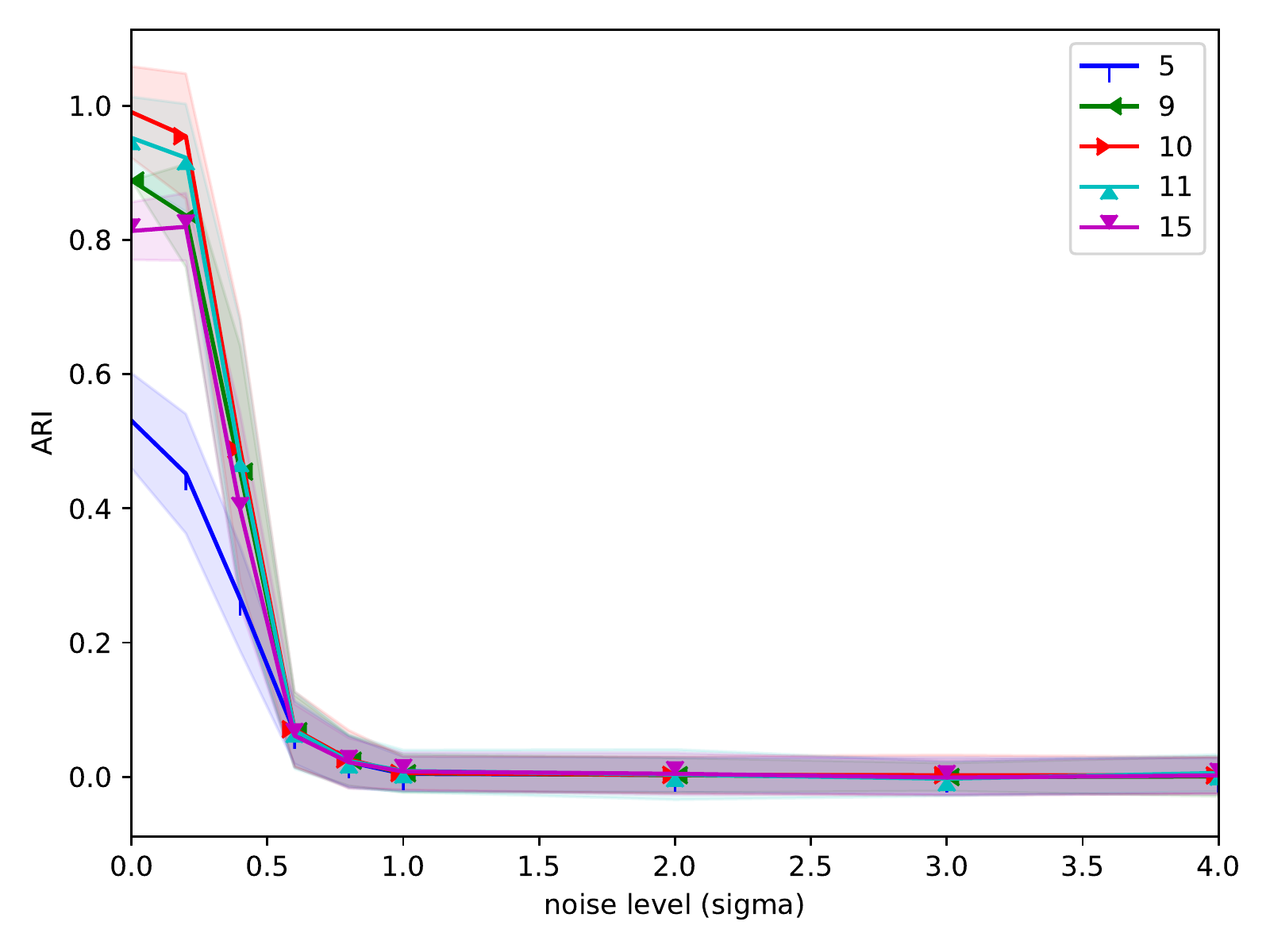}
	\caption{Average and confidence interval for the ARI by different levels of the hyperparameter corresponding to the number of clusters in the cosine setting.}
	\label{fig:ablation-cosine}
\end{figure}

\begin{figure}[h!] 
	\centering
		\includegraphics[width=0.55\textwidth]{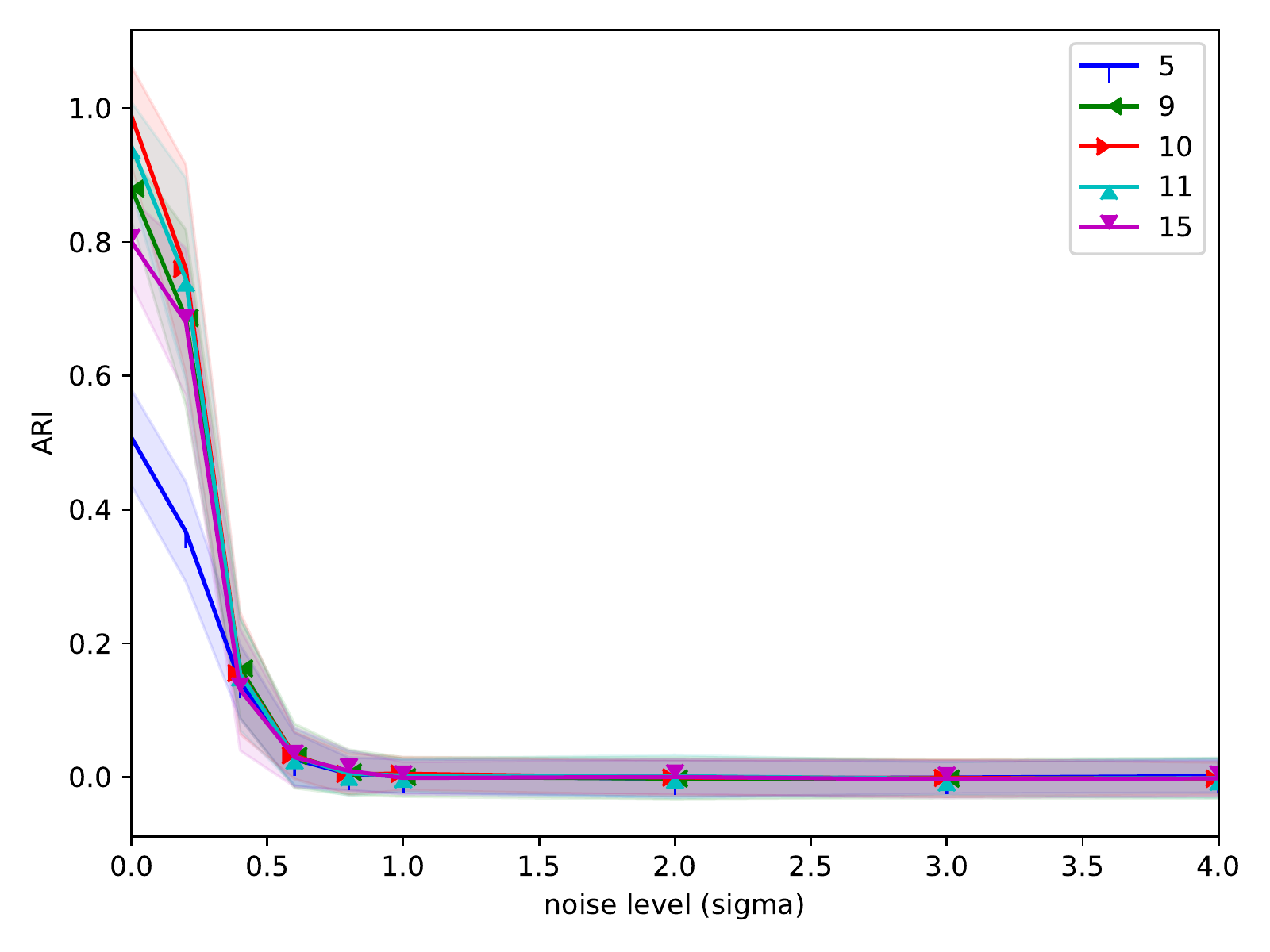}
	\caption{Average and confidence interval for the ARI by different levels of the hyperparameter corresponding to the number of clusters in the Legendre setting.}
	\label{fig:ablation-legendre}
\end{figure}

\begin{figure}[h!] 
	\centering
		\includegraphics[width=0.55\textwidth]{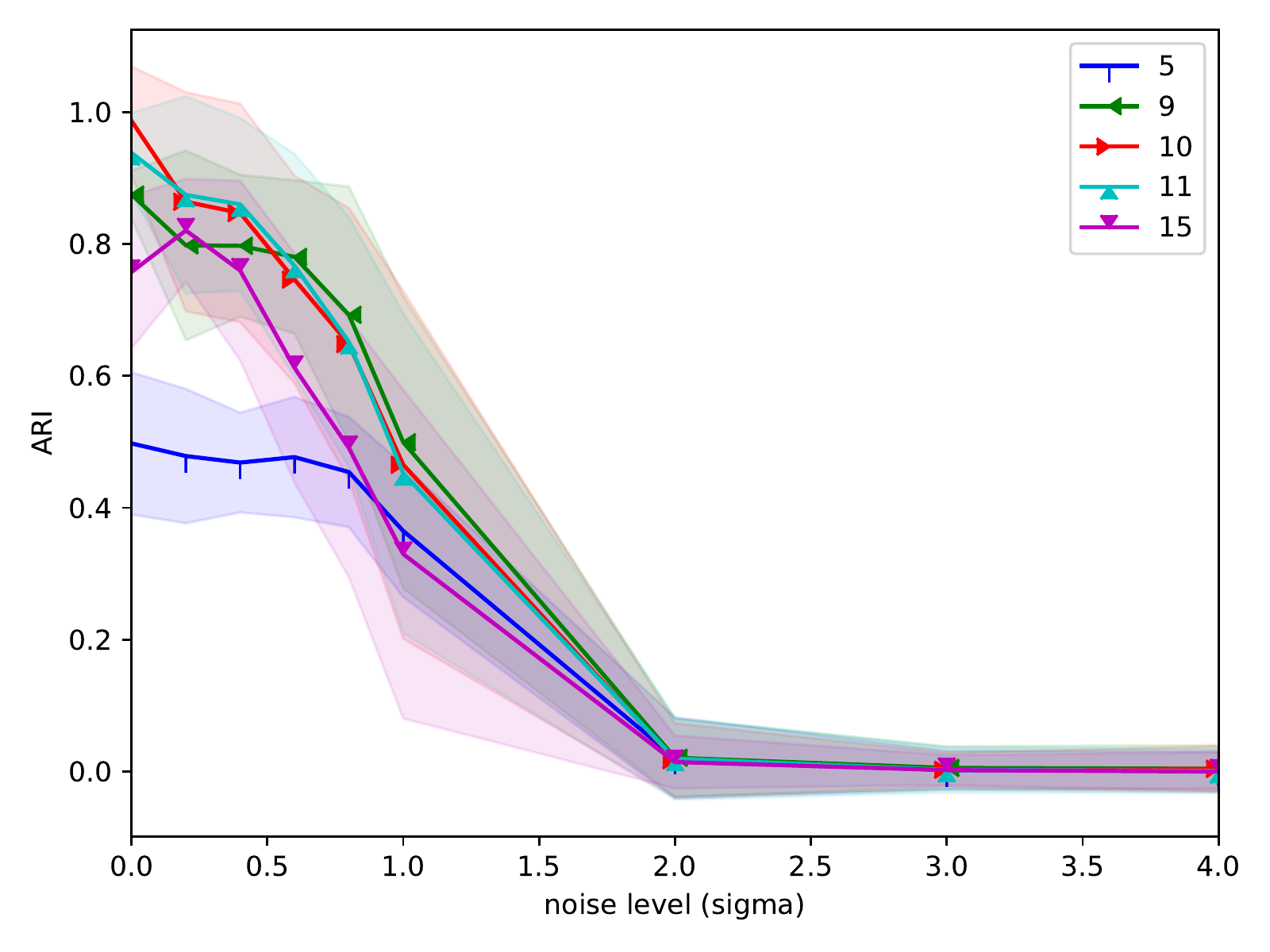}
	\caption{Average and confidence interval for the ARI by different levels of the hyperparameter corresponding to the number of clusters in the Hermite setting.}
	\label{fig:ablation-hermite}
\end{figure}

In figures \ref{fig:ablation-cosine}, \ref{fig:ablation-legendre} and \ref{fig:ablation-hermite} we display the average and confidence interval for the ARI across different hyperparameter levels for the number of clusters used in the clustering component of the pipeline.

\subsection{Real data experiment: time-variation in results}
\label{sec:Appendix - Real data experiment: time-variation in results}

\begin{figure}[h!]
\centering
\begin{minipage}{.48\textwidth}
\centering
	\includegraphics[width=0.99\textwidth]{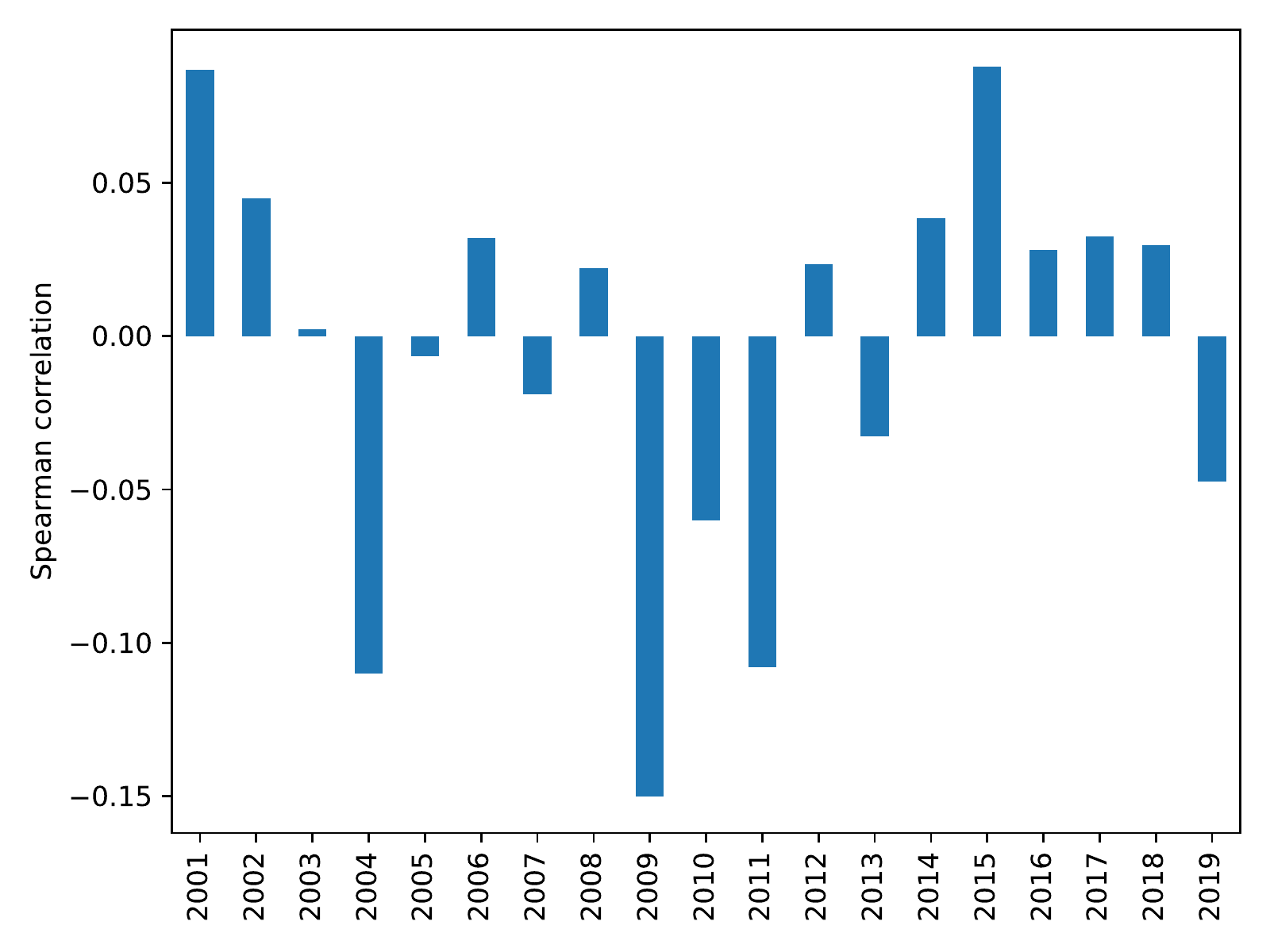}
	\caption{Spearman correlation between the lead-lag matrix row-sums and \textit{average daily trading volume} for each equity, using yearly snapshots of data.}
	\label{fig:rolling_rowsum_average_daily_volume_corr}
\end{minipage}%
\hspace{4mm}
\begin{minipage}{.48\textwidth}
\centering
\includegraphics[width=0.99\textwidth]{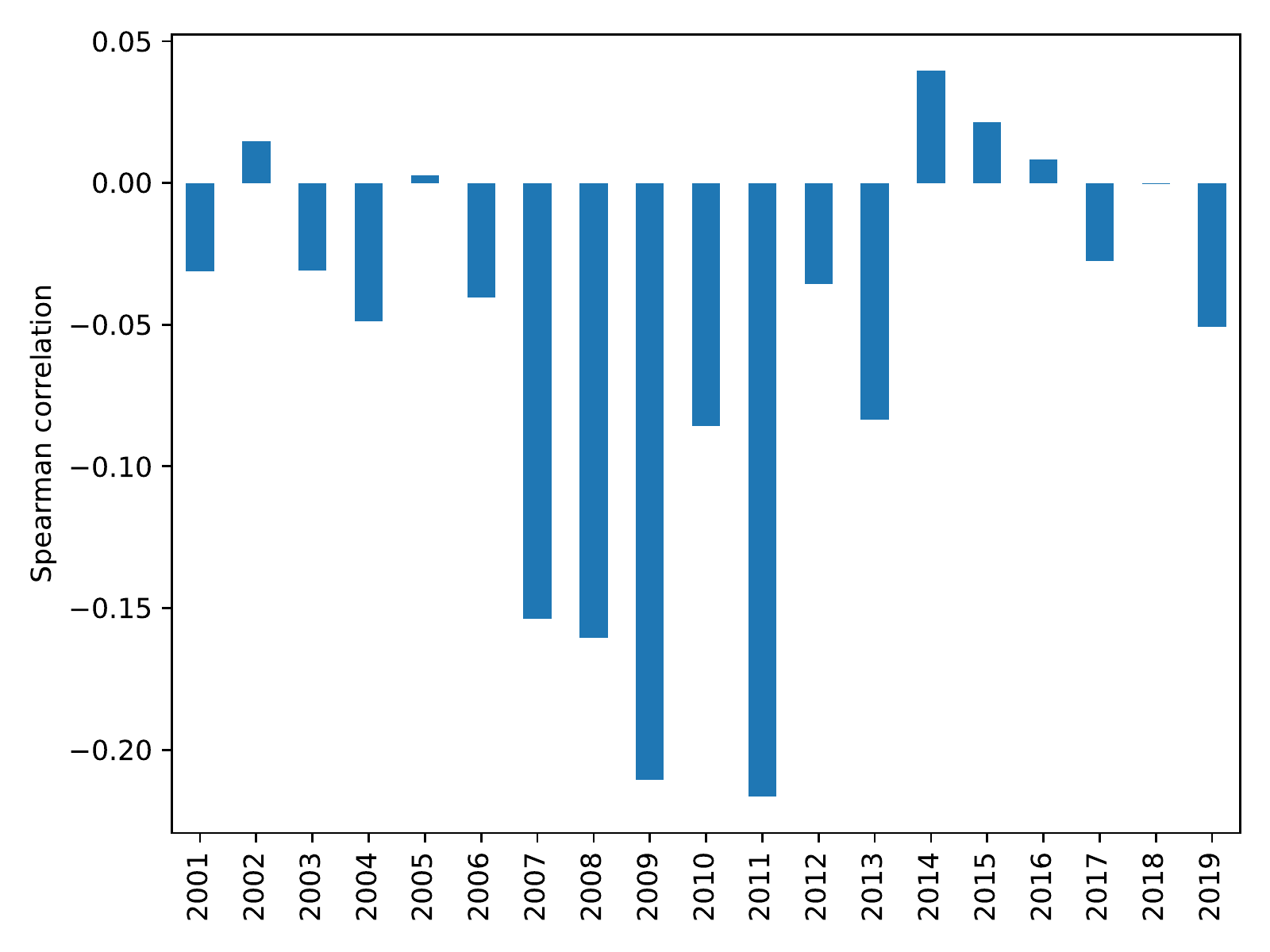}
\caption{Spearman correlation between lead-lag matrix row-sums and \textit{market capitalisation} for each equity, using yearly snapshots of data.}
\label{fig:rolling_rowsum_market_capitalisation_corr}
\end{minipage}
\end{figure}

Figures \ref{fig:rolling_rowsum_average_daily_volume_corr} and \ref{fig:rolling_rowsum_market_capitalisation_corr} display the temporal variation in Spearman correlation between the US equity lead-lag matrix row-sums and a given characteristic (\textit{average daily trading volume} in Figure \ref{fig:rolling_rowsum_average_daily_volume_corr} and \textit{market capitalisation} in Figure \ref{fig:rolling_rowsum_market_capitalisation_corr}) of each equity. We observe that there is substantial temporal variation in each equity's tendency to be a leader (as measured by its lead-lag matrix row-sum) and its underlying characteristic.

\end{document}